\documentclass[preprint,3p]{elsarticle}

\usepackage{amssymb}
\usepackage[table]{xcolor}
\usepackage{amsmath}
\usepackage{amsthm}
\usepackage{booktabs}
\usepackage{algorithm}
\usepackage{algorithmicx}
\usepackage{algpseudocode}
\usepackage{subfigure}
\usepackage{dblfloatfix}
\usepackage{tasks}
\usepackage{soul}

\usepackage{balance}
\usepackage{multirow}

\graphicspath{{images/}}
\newcommand{\eg}{\textit{e}.\textit{g}. }
\newcommand{\algrule}[1][.2pt]{\par\vskip.5\baselineskip\hrule height #1\par\vskip.5\baselineskip}
\definecolor{myblue}{RGB}{0, 0, 0}

\journal{Artificial Intelligence}

\begin{document}

\begin{frontmatter}

\title{Learning a Fast 3D Spectral Approach to Object Segmentation and Tracking over Space and Time}


\author[a,b,d]{Elena Burceanu\corref{cor1}}
\ead{eburceanu@bitdefender.com}

\cortext[cor1]{Corresponding author}

\author[c,d]{Marius~Leordeanu}





\address[a]{Bitdefender, Romania}
\address[b]{University of Bucharest, Romania}
\address[c]{Politehnica University of Bucharest, Romania}
\address[d]{Institute of Mathematics of the Romanian Academy, Romania}

\begin{abstract}
We pose video object segmentation as spectral graph clustering in space and time, with one graph node for each pixel and edges forming local space-time neighborhoods. We claim that the strongest cluster in this video graph represents the salient object. We start by introducing a novel and efficient method based on 3D filtering for approximating the spectral solution, as the principal eigenvector of the graph's adjacency matrix, without explicitly building the matrix. This key property allows us to have a fast parallel implementation on GPU, orders of magnitude faster than classical approaches for computing the eigenvector. Our motivation for a spectral space-time clustering approach, unique in video semantic segmentation literature, is that such clustering is dedicated to preserving object consistency over time, which we evaluate using our novel segmentation consistency measure. Further on, we show how to efficiently learn the solution over multiple input feature channels. Finally, we extend the formulation of our approach beyond the segmentation task, into the realm of object tracking. In extensive experiments we show significant improvements over top methods, as well as over powerful ensembles that combine them, achieving state-of-the-art on multiple benchmarks, both for tracking and segmentation.
\end{abstract}



\end{frontmatter}

\section{Introduction}
\begin{figure*}[!t]
	\begin{center}
		\includegraphics[width=0.95\linewidth]{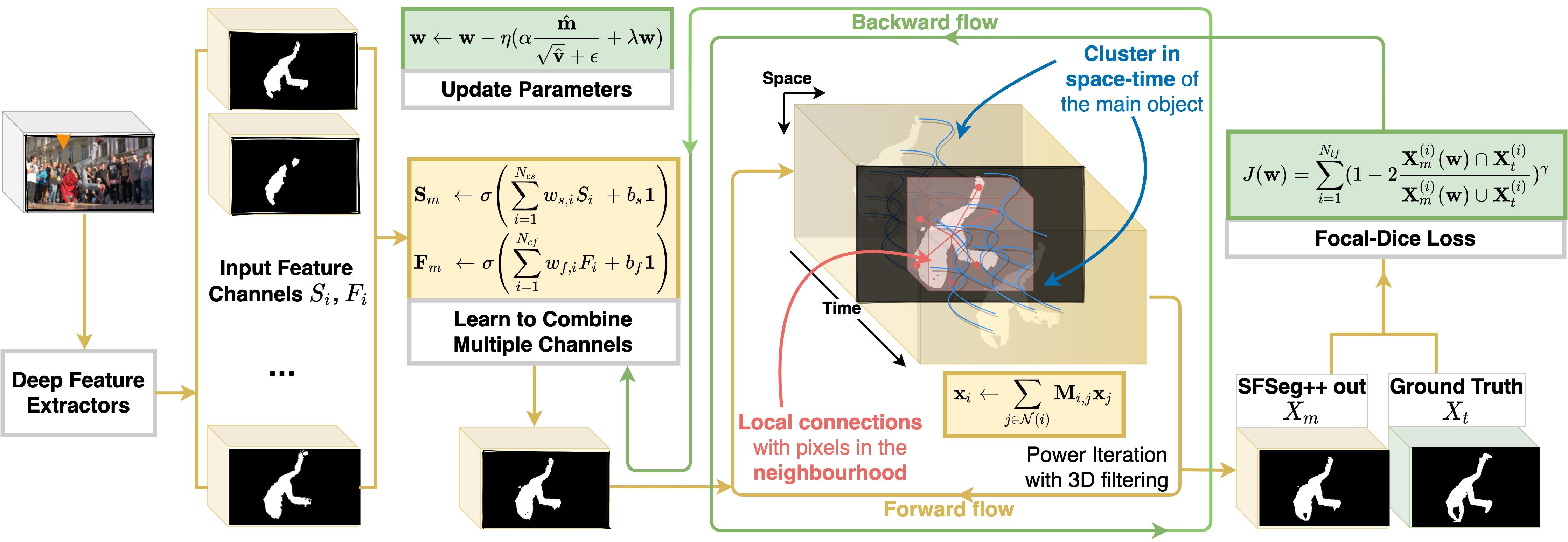}
	\end{center}
	\caption{SFSeg++: Our method for video segmentation learns to combine multiple input feature channels and compute very fast the main eigenvector of the space-time graph with a special set of 3D convolutions. The combination weights for the different channels are learned with gradient descent, by propagating the gradient of the loss through the full spectral algorithm. In experiments, we start from the output of top published segmentation methods and use their combined output volume, in space and time, as graph nodes for our spectral approach, which then computes the final segmentation frame by frame. Please note that this method is general and could learn to combine any type of input feature channels, as tests show. \textbf{SFSeg++ permits full end-to-end learning}, when the deep networks and systems that provide the input feature channels are fully differentiable. Best seen in color.}
	\label{fig: architecture}
\end{figure*}

Elements from a video are interconnected in space and time and have an intrinsic graph structure (Fig.~\ref{fig: architecture}). 
Most existing approaches use higher-level components, such as objects, super-pixels, or features, at a significantly lower resolution. Considering the graph structure in space and time, explicitly at the dense pixel-level, is an extremely expensive problem. Our proposed solution to video object segmentation is based on transforming an expensive eigenvalue problem inspired by spectral clustering, into 3D convolutions on the space-time volume, which makes our algorithm fast, while keeping the properties of spectral clustering. On top of that, and different from the conference version of our paper (\textbf{SFSeg}~\cite{sfseg}) in which we refine the output of a single existing method,
we show how to learn the space-time clustering 
in the context of multiple feature input channels (\textbf{SFSeg++}). This allows learning 
of powerful ensembles in combination with our spectral clustering approach, to obtain
state-of-the-art performance on challenging datasets by a significant margin. We challenge next the rough bounding box shape used for tracking. While they provide a handy way to annotate datasets, bounding boxes are rather imperfect labels since they lead to errors that accumulate, propagate, and are amplified over time. Objects rarely look like boxes, and bounding boxes contain most of the time significant background information or distractors. Since a good object segmentation directly influences tracking performance, we constrain tracking to use such a fine-grained mask in order to reduce the noise transferred from one frame to the next. \textbf{SFSeg++} is at the core of this segmentation, which is seamlessly integrated into a larger tracking system, termed~\textbf{SFTrack++}. 
Our proposed~\textbf{SFTrack++} demonstrates that segmentation and tracking can be naturally linked, as the correct object segmentation is required for robust tracking of objects whose shape is far from being a bounding box.

We are the first, to our best knowledge, to propose an efficient spectral clustering approach, at the core of both object segmentation and tracking in video. Another novelty is that our spectral clustering is also able to adapt and learn the space-time clustering over multiple input feature channels. Most state-of-the-art methods do not use the time constraint, and when they do, they take little advantage of it. Time plays a fundamental factor in how objects move and change in the world, but computer vision does not yet exploit it sufficiently. Consequently, the segmentation outputs of current state-of-the-art methods are not always consistent over time. Our work addresses precisely this aspect and our contribution is demonstrated through solid experiments on multiple benchmarks for segmentation (DAVIS-2016 and SegTrackv2) and tracking (OTB, UAV, NFS, GOT-10k, and TrackingNet), on which we obtain state-of-the-art performance.

This work is a continuation of SFSeg~\cite{sfseg}, in which we demonstrated theoretically and through extensive experiments that the eigenvector of the space-time graph's adjacency matrix is a good solution for salient object segmentation and can be computed fast with a special set of 3D convolutions. We prove theoretically and in practice that SFSeg reaches the same solution as standard routines for eigenvector computation. We also show in experiments that the values in the final eigenvector, with one element per video pixel, confirm the spectral clustering assumption and provide an improved soft-segmentation of the main object. Here we present the full spectral approach for segmentation, SFSeg++. By augmenting SFSeg with learning capability, it learns how to combine multiple input channels in conjunction with the 3D spectral filtering, to significantly improve over the initial version and also over different types of ensembles that combine the output of existing state-of-the-art methods. In continuation, we build on top of SFSeg++ and go beyond the segmentation task, extending the space-time spectral approach to the task of tracking, through the proposed SFTrack++ system. 

One of the key properties of our filtering-based optimization is that it converges to the eigenvector of the graph of pixels in space and time. The segmentation map obtained is spatio-temporally consistent, with a smooth and coherent transition between frames. That is the main reason why the proposed method is best suited for improving or refining the output of any other method, since most existing algorithms do not take full advantage of the space-time consistency. Therefore, when our method is applied on top of the output of another, the noise coming from other objects is removed and missing parts of the object are added back.
Through multiple iterations, the relevant information is propagated step by step to farther away neighbourhoods in space and time, acting as a diffusion. As experiments show, the improvement brought by our algorithm is consistent and reliable - it takes place almost every time and with the same set of hyper-parameters.

\textbf{The main contributions} of this work are the following:

\begin{enumerate}
    \item We formulate the instance segmentation problem in video as an eigenvalue problem on the adjacency matrix of the pixels' graph in space and time. Given this formulation, we provide a fast optimization algorithm, \textbf{SFSeg}, with a special set of 3D filtering operations, which computes the required eigenvector (representing the desired segmentation) without explicitly creating or using the huge graph's adjacency matrix, in which each pixel in space and time has an associated node.
    \item We enrich the initial formulation into a fully learnable approach, over multiple input channels: \textbf{SFSeg++}. In experiments, we show that learning over all segmentation methods available for training on DAVIS~2016, brings in complementary information and significantly improves over the input. SFSeg++ outperforms on the current single best state-of-the-art by $4.2\%$ and the ensemble of top $15$ published methods by $3.1\%$.
    \item We introduce \textbf{TCONT}, a novel temporal consistency metric for segmentation in video, which looks at the temporal continuity of the predictions in video and their stability with respect to the motion field of the object (as estimated by optical flow). The experiments show that our SFSeg++ algorithm is significantly more temporally consistent with respect to ground-truth, when compared to single or ensemble state-of-the-art approaches.
    \item We extend SFSeg++ beyond segmentation, to object tracking, and create \textbf{SFTrack++}, which takes into account the accurate shape of the object vs. the popular, but inaccurate, bounding box approximation. The integration of our special 3D filtering in space and time into object tracking is seamless and efficient, with an immediate boost in performance, as demonstrated on recent challenging tracking benchmarks.
\end{enumerate}

\section{Relation to prior work}
\label{sec: related_work}

Most state-of-the-art methods for video object segmentation use \textbf{CNNs architectures}, pre-trained for object segmentation on other large image datasets. They have a strong image-based backbone and are not designed from scratch with both space and time dimensions in mind. Many solutions~\cite{masktrack, lvo, sfl} adapt image segmentation methods by adding an additional branch to the architecture for incorporating the time axis: motion branch (previous frames or optical flow) or previous masks branch (for mask propagation). Other methods are based on one-shot learning strategies and fine-tune the model on the first video frame, followed by some post-processing refinement~\cite{onavos, osvoss}. Approaches derived from OSVOS~\cite{osvos} do not take the time axis into account. 

\textbf{Improving along the temporal dimension.} Our method better addresses the natural space-time relationship, which is why it is always effective when combined with frame-based segmentation algorithms. The need for time consistency was previously identified. A temporal stability metric was proposed in DAVIS-2016~\cite{davis2016} but in less than a year it was withdrawn \cite{davis-2017}, since the metric was strongly influenced by occlusion, leading to irrelevant comparisons. Temporal stability was based on the Dynamic Time Warping problem, looking to match, between two frames, pixels from the contour of the object, such that the distance between Shape Context Descriptor from the two shapes is minimized. In contrast, our TCONT metric for temporal consistency is based on checking the alignment of nearby predictions using direct and reverse optical flow, and it is not affected by occlusions. Tests in Sec.~\ref{subsec: exp_temporal_consistency} show that our approach brings a complementary value along the time axis, in solving object video segmentation, while improving the temporal consistency of the result. 

\textbf{Relation to different graph representations.} Graph methods are suitable for segmentation and can have different representations, where the \textbf{nodes} can be pixels, super-pixels, voxels, image or video regions \cite{contrastive-random-walk}. While there are works on directed graphs \cite{directed-affinities, directed-groups}, edges are usually undirected, modeled by symmetric similarity functions.
The choice of the representation influences both accuracy and runtime. Specifically, pixel-level representations are computationally extremely expensive, making the problem intractable for high-resolution videos. Our fast solution implicitly uses a pixel-level graph representation: we make a first-order Taylor approximation of the Gaussian kernel (usually used for pairwise affinities) and rewrite it as a sequence of 3D convolutions in the video directly. Thus, we get the desired outcome without explicitly working with the graph. We describe the technical novelties in detail in Sec.~\ref{sec: math_formulation}. 

\textbf{Relation to spectral clustering.} Computing eigenvectors of matrices extracted from data is a classic approach for clustering. There are several choices in the literature for choosing those matrices, the most popular being the Laplacian matrix~\cite{NJW},
normalized~\cite{img_normalized_cut_malik_2000} or unnormalized. Other methods use the random walk matrix~\cite{meila_shi, contrastive-random-walk} or directly the unnormalized adjacency matrix~\cite{marius_iccv2005}. Most methods are based on finding the eigenvectors corresponding to the smallest eigenvalues, while others, including ours, require the leading eigenvectors. \textbf{Graph Cuts} are a popular class of spectral clustering algorithms, with many variants, like normalized~\cite{img_normalized_cut_malik_2000}, average~\cite{cut_average}, min-max~\cite{cut_min_max}, mean cut~\cite{cut_mean} and topological cut~\cite{cut_topological}. 

\textbf{Relation to CRFs.} Discriminative  graphical models such Conditional Random Fields~\cite{lafferty2001conditional} and Discriminative Random Fields~\cite{kumar2003discriminative} are often applied over the segmentation of images and videos (denseCRF~\cite{denseCRF}). Different from the more classical Markov Random Fields (MRFs)~\cite{won1992unsupervised, d-cnn-mrf}, which are generative models, CRFs are more effective as they incorporate the observed data both at the level of nodes as well as edges. Different from our approach, CRFs~\cite{a-obj-flow} have a strict probabilistic interpretation and use inference algorithms (\eg belief propagation, iterative conditional modes, Gibbs sampling, graph-cut) that are significantly more expensive than the simpler eigenvector power iteration that we use for optimizing our non-probabilistic objective score. In experiments, we compare and also combine our method with denseCRF~\cite{denseCRF} and show that the two bring complementary value to the final solution.

\textbf{Relation to image segmentation.} Graph cuts, which are a well-known class of spectral clustering algorithms,
have been widely used in image segmentation~\cite{img_Wu_Leahy, img_normalized_cut_malik_2000}. They are expensive in practice, as they require the computation of eigenvectors of smallest eigenvalues for very large Laplacian matrices. Fast graph-based algorithm for image segmentation exist, such as \cite{img_efficient_graph_based_felzen_2004}, which is linear in the number of edges. It is based on a heuristic for building the minimum spanning tree and is still used as a starting point in current methods. Another approach~\cite{img_weakly_superv_2015} is to learn image regions with spectral graph partitioning and solve segmentation with convex optimization.

\textbf{Relation to video segmentation.} Many video segmentation methods adapt existing segmentation approaches, as it is the parametric graph partitioning model over superpixels~\cite{video_efficient_param_graph_part_2015}. Hierarchical graph-based segmentation over RGBD video sequences \cite{video_efficient_hierarch_rgbd_2018} also groups pixels into regions. The problem is solved using bipartite graph matching and minimizing the spanning tree. Another efficient graph cut method~\cite{video_unsup_segm_motion_2018} is applied on a subset of pixels. Different from ours, all efficient methods group pixels into superpixels, grid regions, or object proposals~\cite{nlc, fcopvs, bilateral, dondera} to handle the computational burden. However, the hard initial grouping of pixels could carry initial errors into the final solution, as it could miss details available only at the original pixel resolution.

\textbf{Relation to space-time segmentation.} When focusing on the spatio-temporal property, there are several graph-based probabilistic approaches modeling Markov Random Fields~\cite{d-cnn-mrf} or more specifically, CRFs~\cite{a-obj-flow}. As detailed in a previous paragraph, additional to modeling the problem, our novelty stays in making the formulation efficient and tractable. Other solutions are based on deep models starting from classical end-to-end trainable deep learning pipeline~\cite{e-stemseg}, some keeping and querying a memory of past frames~\cite{b-mem-networks}, while others have different branches for temporal and spatial dimensions~\cite{c-stcnn}, treating them separately. Compared with all those works, our approach combines the empirical learning-based method with a classical mathematical formulation. We propose an efficient power iteration algorithm for clustering foreground vs background in the video's volume of pixels.

\textbf{Relation to ensembles.} 
Ensembles were widely used over time for improving the single model performance They better capture the underlying data distribution, being statistically proven to be more robust outliers and uncorrelated errors. Usually, ensembles gets the first places in competitions~\cite{competition_neurips, workshop_cvpr}. Fusion over multiple existing models ranges from random trees and bagging to learning weights for each component~\cite{weight_ensemble}. Closer to the segmentation and tracking tasks in video, weighted ensemble solutions were used over bounding boxes for detecting objects~\cite{weighted_bbox_ensemble} or over-dense medical segmentation predictions~\cite{weight_ensemble_medical_segm}. In our implementation, we learn to combine multiple methods, passing the ensemble result through the spectral formulation, which is also part of the gradient flow.

\textbf{Relation between video segmentation and object tracking.} With a few notable exceptions \cite{segtrack, mots}, most tracking solutions use internally hidden layer representations extracted from the previous frame's rough bounding box prediction \cite{kys, prdimp, atom, siamfc++}, rather than a fine-grained segmentation mask, as in our proposed approach. Also, most of them do not take into account multiple perceptions for the input frame and operate over a unique feature extractor~\cite{atom, ocean, siamfc}. There are a few trackers though that combine two models for adapting to sudden changes while remaining robust to background noise, by explicitly modeling the different pathways \cite{kys, multi-path-tracking}. In contrast, we show that SFTrack++, our end-to-end multi-channel approach, learns over $5$ input channels, coming from different models.
\color{black}

\textbf Our work extends to the space-time domain the goals of classical spectral clustering approaches~\cite{marius_iccv2005, meila_shi}, by showing that a special set of 3D convolutions can effectively perform clustering in space and time. The final solution becomes the leading eigenvector of an adjacency matrix, computed fast and stable with power iteration, as shown in Sec.~\ref{sec: math_formulation}. Note that the unnormalized adjacency matrix in combination with power iteration is the least expensive spectral approach possible - the only one that can be factored into fast 3D convolutions. This possibility gives our method efficiency and speed (Sec.~\ref{sec: algorithm}). 

\section{Problem Formulation}
\label{sec: math_formulation}

We formulate salient object segmentation in video as a graph partitioning problem (foreground vs background), where the graph is both spatial and temporal. Each node $i$ represents a pixel in the space-time volume, which has $N = N_f \times H \times W$ pixels. $N_f$ is the number of frames and $(H,W)$ the frame size. Each edge captures the similarity between two pixels and is defined by the pairwise function $\mathbf{M}_{i, j}$. We require the pairwise connections between pixels $i$ and $j$, in space and time, to be symmetric and non-negative, defining a $N \times N$ adjacency matrix $\mathbf{M}$. We take into account only the local connections in space-time, so $\mathbf{M}$ is sparse. 

Let $\mathbf{s}$ and $\mathbf{f}$ be feature vectors of size $N \times 1$ with a feature value for each node. They will be used in defining the similarity function $\mathbf{M}_{ij}$ (Eq.~\ref{eq: mij_equation}).
For now, we consider the simplest case when $(\mathbf{s}_i,\mathbf{f}_i)$ represent single-channel features (\eg they could be soft masks, grey level values, edge or motion cues, or any pre-trained features). Later on, we show how we can easily adapt the formulation to the multi-channel feature case. We define the edge similarity $\mathbf{M_{i, j}}$ using a Gaussian kernel:

\begin{equation}
    \begin{aligned}
        \mathbf{M}_{i,j} &= \mathbf{s}_i^p \mathbf{s}_j^p e^{- \alpha ( \mathbf{f}_i -  \mathbf{f}_j)^2 - \beta \mathbf{dist}^2_{i, j}} \\
        &= \mathbf{s}_i^p \mathbf{s}_j^p e^{- \alpha ( \mathbf{f}_i -  \mathbf{f}_j)^2} \mathbf{G}_{i, j}
    \end{aligned}
	\label{eq: mij_equation}
\end{equation}

\begin{equation}
    \begin{aligned}
    	 \mathbf{M}_{i,j} 
    	 & \approx \underbrace{\mathbf{s}_i^p \mathbf{s}_j^p}_{\textnormal{unary terms}} \underbrace{[1 - \alpha (\mathbf{f}_i -  \mathbf{f}_j)^2] \mathbf{G}_{i, j}}_{\textnormal{pairwise terms}}.
    \end{aligned}
	\label{eq: mij_equation_approx}
\end{equation}
In graph methods, it is common to use two types of terms for representing the model over the graph. Unary terms are about individual node properties, while pairwise terms describe relations between pairs of nodes. In our case, $\mathbf{s}_i$, $\mathbf{s}_j$ describe individual node properties, whereas $\mathbf{f}_i$, $\mathbf{f}_j$ are used to define the pairwise similarity kernel between the two nodes. Note that in Eq.~\ref{eq: mij_equation_approx} we approximate the Gaussian kernel with its first-order Taylor expansion. The approximation is crucial in making the filtering approach possible, as shown next. Hyper-parameters $p$ and $\alpha$ control the importance of those terms.

To partition the space-time graph of video pixels, we want to find the strongest cluster in this graph. We first represent a segmentation solution (i.e., cluster in the space-time graph) with an indicator vector $\mathbf{x}$, that has one element for each node in the 3D space-time volume, such that $x_i=1$ if node (pixel) $i$ is in the video segmentation cluster (foreground) and  $x_i=0$ otherwise (background). We define the clustering score to be the sum over all pairwise similarity terms $\mathbf{M}_{ij}$ between the nodes inside the cluster. The higher this score, the stronger the sum of connections and the cluster. The segmentation score can be written compactly in matrix form as $S(\mathbf{x})=\mathbf{x}^\top \mathbf{M}\mathbf{x}$. Similar to other spectral approaches in graph matching \cite{marius_iccv2005}, we find the segmentation solution $\mathbf{x}_s$ that maximizes $S(\mathbf{x})$ under the relaxed constraints $\|\mathbf{x}\|_2 = 1$. Fixing the L2 norm of $\mathbf{x}$ is needed since only relative soft segmentation values matter. Thus, the optimization problem becomes
the maximization of the Raleigh quotient:

\begin{equation}
    \begin{aligned}
        \mathbf{x}_s = \underset{\mathbf{x}}{\mathrm{argmax}}  (\mathbf{x}^\top \mathbf{M} \mathbf{x } / \mathbf{x}^\top \mathbf{x}).
    \end{aligned}
	\label{eq: argmax_eq}
\end{equation}

The global optimum solution is the principal eigenvector of $\mathbf{M}$. $\mathbf{M}$ is symmetric and has non-negative values, so the solution will also have non-negative elements, by Perron-Frobenius theorem~\cite{frobenius}. The final segmentation could be simply obtained by thresholding. However, matrix $\mathbf{M}$, even for a small video has 20 million nodes (50 frames of $480 \times 854$), making the problem of finding the leading eigenvector with standard procedures intractable (Sec~\ref{subsec: computational_complexity}).

Next, we show how to take advantage of the first-order expansion of the pairwise terms defining $\mathbf{M}$ and break power iteration into several very fast 3D convolutions in space and time, directly on the feature maps, without explicitly using the huge adjacency matrix of the video. Our method receives as input pixel-level feature maps and returns a final segmentation, as the solution $\mathbf{x}_s$ to Eq.~\ref{eq: argmax_eq}.

\subsection{Power iteration with pixel-wise iterations}
We apply power iteration algorithm to compute the eigenvector. At iteration $k+1$, we have Eq.~\ref{eq: power_it_eq}:
\begin{equation}
	\mathbf{x}^{k+1}_i \leftarrow \sum_{j \in \mathcal{N}(i)}\mathbf{M}_{i, j} \mathbf{x}^k_j, 
	\label{eq: power_it_eq}
\end{equation}
where, after each iteration, the solution is normalized to unit norm and $\mathcal{N}(i)$ is the set of neighbors pixels with $i$, in space and time. Expanding $\mathbf{M}_{i, j}$ (Eq.~\ref{eq: mij_equation_approx}), Eq.~\ref{eq: power_it_eq} becomes:

\begin{equation}
\mathbf{x}^{k+1}_i \leftarrow \alpha \mathbf{s}_i^p \sum_{j \in \mathcal{N}(i)} \mathbf{s}_j^p [\alpha^{-1} - \mathbf{f}_i^2 - \mathbf{f}_j^2 + 2 \mathbf{f}_i \mathbf{f}_j] \mathbf{G}_{i, j} \mathbf{x}^k_j,
\end{equation}

\begin{equation}
\label{eq: before_matrix_form}
  \begin{aligned}
    \mathbf{x}^{k+1}_i \leftarrow  \alpha \mathbf{s}_i^p ( \alpha^{-1} - \mathbf{f}_i^2) \sum_{j \in \mathcal{N}(i)} \mathbf{s}_j^p  \mathbf{G}_{i, j} \mathbf{x}^k_j -\\
    \alpha \mathbf{s}_i^p \sum_{j \in \mathcal{N}(i)}  \mathbf{s}_j^p \mathbf{f}_j^2 \mathbf{G}_{i, j} \mathbf{x}^k_j \\
    2 \alpha \mathbf{s}_i^p \mathbf{f}_i  \sum_{j \in \mathcal{N}(i)} \mathbf{s}_j^p \mathbf{f}_j \mathbf{G}_{i, j} \mathbf{x}^k_j.
  \end{aligned}
\end{equation}

\subsection{Power iteration using 3D convolutions}
In Eq.~\ref{eq: before_matrix_form} we observe that the links between the nodes are local ($\mathbf{M}$ is sparse) and we can replace the sums over neighbours with local 3D convolutions in space and time. Thus, we rewrite Eq.~\ref{eq: before_matrix_form} as a sum of convolutions in 3D:

\begin{equation}
\begin{aligned}
\mathbf{X}^{(crt)} \leftarrow \mathbf{S}^p \cdot (\alpha^{-1} \mathbf{1} - \mathbf{F}^2) \cdot G_{3D} * (\mathbf{S}^p  \cdot \mathbf{X}^k) - \\
                        \mathbf{S}^p \cdot G_{3D} * (\mathbf{F}^2 \cdot \mathbf{S}^p \cdot \mathbf{X}^k) + \\
                        2 \mathbf{S}^p \cdot \mathbf{F}\cdot G_{3D} * (\mathbf{F} \cdot \mathbf{S}^p \cdot \mathbf{X}^k),
\end{aligned}
\label{eq: matrix_full_eigenvector}
\end{equation}

\begin{equation}
\label{eq: sfseg_normalize}
\mathbf{X}^{k+1} \leftarrow \mathbf{X}^{(crt)} / \|\mathbf{X}^{(crt)}\|_2,
\end{equation}

where $*$ is a convolution over a 3D space-time volume with a 3D Gaussian filter ($G_{3D}$), $\cdot$ is an element-wise multiplication, 3D matrices $\mathbf{X}^k, \mathbf{S}, \mathbf{F}$ have the original video shape ($N_f \times H \times W$) and $\mathbf{1}$ is a 3D matrix with all values 1. We transformed the 
standard form of power iteration in Eq.~\ref{eq: power_it_eq} in several very fast matrix operations: 3 convolutions and 13 element-wise matrix operations (multiplications and additions), which are local operations that can be parallelized. 

\subsection{Multi-channel SFSeg++}
Our approach in Eq.~\ref{eq: matrix_full_eigenvector} can easily accommodate multiple feature channels. We extend the initial SFSeg formulation such that it can learn how to combine the input feature maps for improving its final result into multi-channel versions of the unary and pairwise maps: $\mathbf{S}_m$ and $\mathbf{F}_m$, respectively, as described in the next Eq.~\ref{eq: sfseg++_1}. 
\begin{equation}
    \begin{aligned} 
        \mathbf{S}_m \leftarrow \sigma(\sum_{i=1}^{N_{cs}} w_{s, i} \mathbf{S}_i + b_s \mathbf{1}),\\
        \mathbf{F}_m \leftarrow  \sigma(\sum_{i=1}^{N_{cf}} w_{f, i} \mathbf{F}_i  + b_f \mathbf{1}),
    \end{aligned}
\label{eq: sfseg++_1}
\end{equation}
where $\sigma$ is the sigmoid function $\sigma(x) = 1/(1 + e^x)$, $N_{cs}$ and $N_{cf}$ are the number of unary and pairwise input feature channels, respectively, $\mathbf{1}$ is an all-one matrix for the bias terms and $w_{s, i}, w_{f, i}, b_s, b_f$ are their corresponding learned weights.

The final power iteration algorithm with 3D filtering for the multi-channel case is described in the next equations: Eq.~\ref{eq: sfseg++_2} and Eq.~\ref{eq: sfseg++_normalize}, which follow immediately from the single-channel Eq.~\ref{eq: matrix_full_eigenvector} and Eq.~\ref{eq: sfseg_normalize}:

\begin{equation}
    \begin{aligned} 
        \mathbf{X}_m^{(crt)} \leftarrow \mathbf{S}_m^p \cdot (\alpha^{-1} \mathbf{1} - \mathbf{F}_m^2) \cdot G_{3D} * (\mathbf{S}_m^p  \cdot \mathbf{X}_m^k) - \\
                    \mathbf{S}_m^p \cdot G_{3D} * (\mathbf{F}_m^2 \cdot \mathbf{S}_m^p \cdot \mathbf{X}_m^k) + \\
                     2 \mathbf{S}_m^p \cdot \mathbf{F}_m \cdot G_{3D} * (\mathbf{F}_m \cdot \mathbf{S}_m^p \cdot \mathbf{X}_m^k),
    \end{aligned}
\label{eq: sfseg++_2}
\end{equation}

\begin{equation}
\label{eq: sfseg++_normalize}
\mathbf{X}_{m}^{k+1} \leftarrow \mathbf{X}^{(crt)}_{m} / \|\mathbf{X}^{(crt)}_{m}\|_2.
\end{equation}

\subsection{Learning over Multi-channel SFSeg++}
\label{subsec: learning}
We learn $w_{s, i}, w_{f, i}, b_s, b_f$ parameters by minimizing the Focal-Dice loss~\cite{focal} over the training video frames, proved to be suitable for segmentation tasks (Eq.~\ref{eq: update_w}):
\begin{equation}
\begin{aligned}
J(\mathbf{w}_t) = \sum_{i=1}^{N_{tf}}(1 - 2\frac{\mathbf{X}^{(i)}_m (\mathbf{w}_t) \cap \mathbf{X}^{(i)}_t}{\mathbf{X}^{(i)}_{m}(\mathbf{w}_t) \cup \mathbf{X}^{(i)}_t}) ^ \gamma,
\end{aligned}
\label{eq: update_w}
\end{equation}
where $N_{tf}$ is the number of training frames and $X_t$ is the ground-truth video segmentation.

We optimize the loss using a state-of-the-art gradient descent based optimizer (AdamW \cite{adamw-amsgrad}):
\begin{equation}
\begin{aligned}
\mathbf{w}_t \leftarrow \mathbf{w}_{t-1} - \eta_t (\alpha \frac{\hat{\mathbf{m}}_t}{\sqrt{\hat{\mathbf{v}}_t} + \epsilon} + \lambda \mathbf{w}_{t-1}),
\end{aligned}
\end{equation}
where $\eta_t$ is the scaling factor and  $\nabla_{\mathbf{w}} J(\mathbf{w}_{t-1})$ is used for updating first and second momentum vectors: $\hat{\mathbf{m}}_t$ and $\hat{\mathbf{v}}_t$ respectively, at optimization step $t$.
 
We make SFSeg++ trainable by using a soft-binarization between iterations. Consequently, it can learn end-to-end, from the original input frames all the way to final output, in the case when the input feature channels are also learnable end-to-end.

\subsection{TCONT: Temporal consistency metric}
\label{subsec: temporal_consistency}

We introduce TCONT, a new metric that measures the consistency in time of a video segmentation input. Different from other metrics in segmentation, this is specifically thought for video segmentation
and focuses on the key distinction between a video seen as a whole and a video seen as just a set of frames. We first observe that a certain segmentation method could perform well at the level of individual frames, without being able to provide the desired consistency and coherency of the object shape over time. This aspect is important for video object segmentation methods and should constitute a complementary dimension for measuring performance, different from the standard, per frame, IoU. Ideally, one wants a segmentation procedure to be both accurate per frame as well as consistent in time. As we see in tests, while the two aspects are correlated, they are not the same.

Here we introduce the TCONT metric, which rebuilds each frame in the segmentation video by taking into account its shape continuity from the left and right frames, via optical flow ($\overrightarrow{\mathbf{OF}_{dir}}, \overleftarrow{\mathbf{OF}_{rev}}$). Ideally, if we warp the object masks from the left and right frames into the current frame, using optical flow, we would want to obtain the same shape, also identical to the current shape in the middle. In order to evaluate this desired property, we transform the segmentations of the previous and next frames into the current one, using optical flow, and then add all three segmentations, past, present and future, with equal proportions. This ensemble of three segmentations, which produce a combined mask, as the average of the three
can now be evaluated on its own, with respect to ground-truth or with respect to the initial current, per frame segmentation.

Intuitively, a large TCONT agreement score means that the frames are contiguous and movement-consistent, such that the segmentation correctly changes over time according to object movements in the video, as expressed by optical flow (without wrongly having large parts of the segmentation flickering in noisy ways). If such time consistency exists and the initial masks are close to ground-truth in similar ways, it is expected that the average mask of the three will maintain its high IoU measure with respect to ground-truth. However, if the segmentations at the three moments in time are not consistent temporally with respect to the object motion field, as estimated by the optical flow, then we could expect the average of the three to produce a degraded, blurry, less accurate mask.

Mathematically we define the TCONT metric as follows:
\begin{equation}
    \begin{aligned}
        \mathbf{x}_{tcont} &= \frac{1}{3} (\overrightarrow{\mathbf{OF}_{dir}}(\mathbf{x}_{1:n-1}) + \mathbf{x}_{2:n-1} + \overleftarrow{\mathbf{OF}_{rev}}(\mathbf{x}_{2:n}))\\
    \end{aligned}
	\label{eq: temp_consist_output}
\end{equation}

\begin{equation}
    \begin{aligned}
        TCONT_{gt} &= IoU(\mathbf{x}_{tcont}, \mathbf{gt}_{2:n-1}),
    \end{aligned}
	\label{eq: temp_consist_metric}
\end{equation}

where $\mathbf{x}$ is the soft video segmentation, $\mathbf{gt}$ is the ground-truth segmentation to which we relate, $n$ is the number of frames, $IoU$ is the IoU metric computed frame by frame, using a certain threshold, and $\overrightarrow{\mathbf{OF}_{dir}}, \overleftarrow{\mathbf{OF}_{rev}}$ represents the direct and reverse optical flow respectively. In Sec.~\ref{subsec: exp_temporal_consistency} we validate experimentally that SFSeg++ algorithm significantly improves the temporal consistency of the initial input segmentation, while also improving the overall IoU performance. This result is confirmed in most of the
experimental setups in Sec.~\ref{sec: experiments}.

\section{Algorithm}
\label{sec: algorithm}
Next, we present our algorithm, which applies to both single or multiple input channels. SFSeg++ (Alg.~\ref{alg: power_iteration}) starts by combining the input channels $\mathbf{S}_i$, $\mathbf{F}_i$, which could be of any kind: lower-level (optical flow, edges, gray-level values) or higher-level pre-trained semantic features (deep features or initial soft/hard segmentation maps) into the new feature maps $\mathbf{S}_m$ and $\mathbf{F}_m$. In the multi-channel SFSeg++ experiments we use as input channels the segmentation output of 15 top methods in the literature. 

Next, we initialize the solution $\mathbf{X}_m$ either with a uniform vector or with the segmentation provided by $\mathbf{S}_m$.
At each iteration, we first select a time frame around the current one. In Step 2 we multiply the corresponding matrices, apply the convolutions, compose the results and obtain the new segmentation mask for pixels in the current frame, using the space-time operations (as in Eq.~\ref{eq: matrix_full_eigenvector}). At evaluation time, the solution needs to be binary, so after each iteration, we project the solution in a more discrete space (see Sec.~\ref{subsec: binarization}). 
 
\begin{algorithm}[t!]
    \caption{SFSeg++: We linearly combine input channels into one single-channel, using a set of weights (that we learn), then pass it through a pixel-wise sigmoid function. At each iteration, we pass through the whole video and compute the updated soft-segmentation $\mathbf{X}_m$. In the first step, we select a time window around current frame $[i-t, i+t]$ ($t=6$ in experiments). Next, we compute the eigenvector using the proposed 3D convolutions. At the end of the continuous power iteration (for $N_{cont}$ iterations), we start soft-binarizing the solution at the end of each iteration until all $N_{iter}$ iterations are completed (see Sec.~\ref{subsec: binarization} for details).}
    \label{alg: power_iteration}
    $\mathbf{S}_i, \mathbf{F}_i$ - \textbf{Input}: unary and pairwise video feature maps\\
    $N_f$ \,\,\,\,\,\,\,- \textbf{Input}: number of video frames \\
    $N_{iter}$ \,\,- \textbf{Input}: total number of iterations \\
    $N_{cont}$ \,- \textbf{Input}: number of continuous space iterations \\
    $\mathbf{X}_m$ \;\;\;\,- \textbf{Output}: salient object segmentation in video
    \begin{algorithmic}[1]
        \algrule
        \Statex $\triangleright$ \textbf{Initialization.} Note that for single-channel case, we directly assign maps to $\mathbf{S}_m$ and $\mathbf{F}_m$, without the sigmoid:
        \State $\mathbf{S}_m \leftarrow \sigma(\sum_{i=1}^{N_{cs}} w_{s, i} \mathbf{S}_i + b_s \mathbf{1})$
        \State $\mathbf{F}_m \leftarrow  \sigma(\sum_{i=1}^{N_{cf}} w_{f, i} \mathbf{F}_i  + b_f \mathbf{1})$
        \State $ \mathbf{X}_m \leftarrow \mathbf{S}_m$\newline

        \For{$iter$ {\bfseries in} $[1.. N_{iter}]$}
            \For{$i$ {\bfseries in} $[(t+1).. (N_f-t)]$}

                $\triangleright$ \textbf{STEP 1:} Temporal window around frame $i$:
                \State $\mathbf{X}_{(t)}, \mathbf{S}_{(t)}, \mathbf{F}_{(t)} \leftarrow T_{i, t}(\mathbf{X}_m, \mathbf{S}_m, \mathbf{F}_m) [i-t:i+t]$\newline
                
                $\triangleright$ \textbf{STEP 2:} Compute new segmentation:
                \State $\mathbf{T1} \leftarrow  (\alpha^{-1}\mathbf{1} - \mathbf{F}_{(t)}^2) \cdot G_{3D} * (\mathbf{S}^p_{(t)} \cdot \mathbf{X}_{(t)}) $
                \State $\mathbf{T2} \leftarrow - G_{3D} * (\mathbf{F}_{(t)}^2 \cdot \mathbf{S}_{(t)}^p \cdot \mathbf{X}_{(t)})$
                \State $\mathbf{T3} \leftarrow 2 \mathbf{F}_{(t)} \cdot G_{3D} * (\mathbf{F}_{(t)} \cdot \mathbf{S}_{(t)}^p \cdot \mathbf{X}_{(t)})$\newline
                \State $\mathbf{X}_{new}[i] \leftarrow \mathbf{S}_{(t)}^p \cdot (\mathbf{T1} + \mathbf{T2} + \mathbf{T3}) $
            \EndFor
        \State $\mathbf{X}_m \leftarrow$ normalize$(\mathbf{X}_{new})$\newline
        
        $\triangleright$ \textbf{STEP 3:} Soft-binarization (using a sigmoid):
        \If{$iter > N_{cont}$}
            \State $\mathbf{X}_m \leftarrow \sigma(\mathbf{X}_m)$
        \EndIf
        \EndFor
    \end{algorithmic}
\end{algorithm}

\subsection{Soft-binarization: from spectral continuous towards the final discrete solution}
\label{subsec: binarization}
The spectral solution over the power iterations is continuous. However, at the very end, we need a binary, hard segmentation map for the object of interest. While the relaxed, continuous solution is globally optimal w.r.t Eq.~\ref{eq: argmax_eq}, there is no guarantee that a simple, binary thresholding of the final continuous solution will retain optimality. In fact, as previously observed in the graph matching literature such optimality is often lost by thresholding, so keeping the continuous solution as close as possible to the initial discrete domain comes with a better final performance~\cite{ipfp}, even though the global optimum in the spectral space is affected. Therefore, we took a similar approach, and after the continuous power iteration is over, we continue with a set of iterations after each of which the solution is soft-binarized, with a sigmoid function (controlled by a parameter) that gradually brings the solution to the discrete domain. Thus, after each soft-binarization iteration, the solution is projected onto an almost discrete space. So, after the very last iteration, we apply a hard threshold on a solution that is much closer to the desired discrete space than the continuous spectral solution.

\subsection{Computational complexity}
\label{subsec: computational_complexity}
We compare the standard power iteration eigenvector computation with our filtering formulation, both from qualitative and quantitative points of view. In terms of the quantitative comparative analysis, we will look at both accuracy (how close the filtering is to the exact solution) and speedup (how fast the 3D filtering approach is to the classic Lanczos method for computing the principal eigenvector of a matrix).

Lanczos~\cite{lanczos_1950} method for sparse matrices has $\mathcal{O}(k N_f N_p N_i)$ complexity for computing the leading eigenvector,
where $k$ is the number of neighbours for each node, $N_f$ the number of frames in video, $N_p$ the number of pixels per frame and $N_i$ the number of iterations. Our full iteration algorithm has also $\mathcal{O}(k N_f N_p N_i)$ complexity, but with highly parallelizable operations, comparing to Lanczos.  The Gaussian filters are separable, so the 3D convolutions can be broken into a sequence of three vector-wise convolutions, reducing the complexity $\mathcal{O}(k)$ for filtering to $3 \mathcal{O}(k^{\frac{1}{3}})$: $3$*$7$*$7$=$147$ vs $3$+$7$+$7$=$17$ for a $3$x$7$x$7$ kernel.

\begin{figure}[t]
	\begin{center}
		\includegraphics[width=0.88\linewidth]{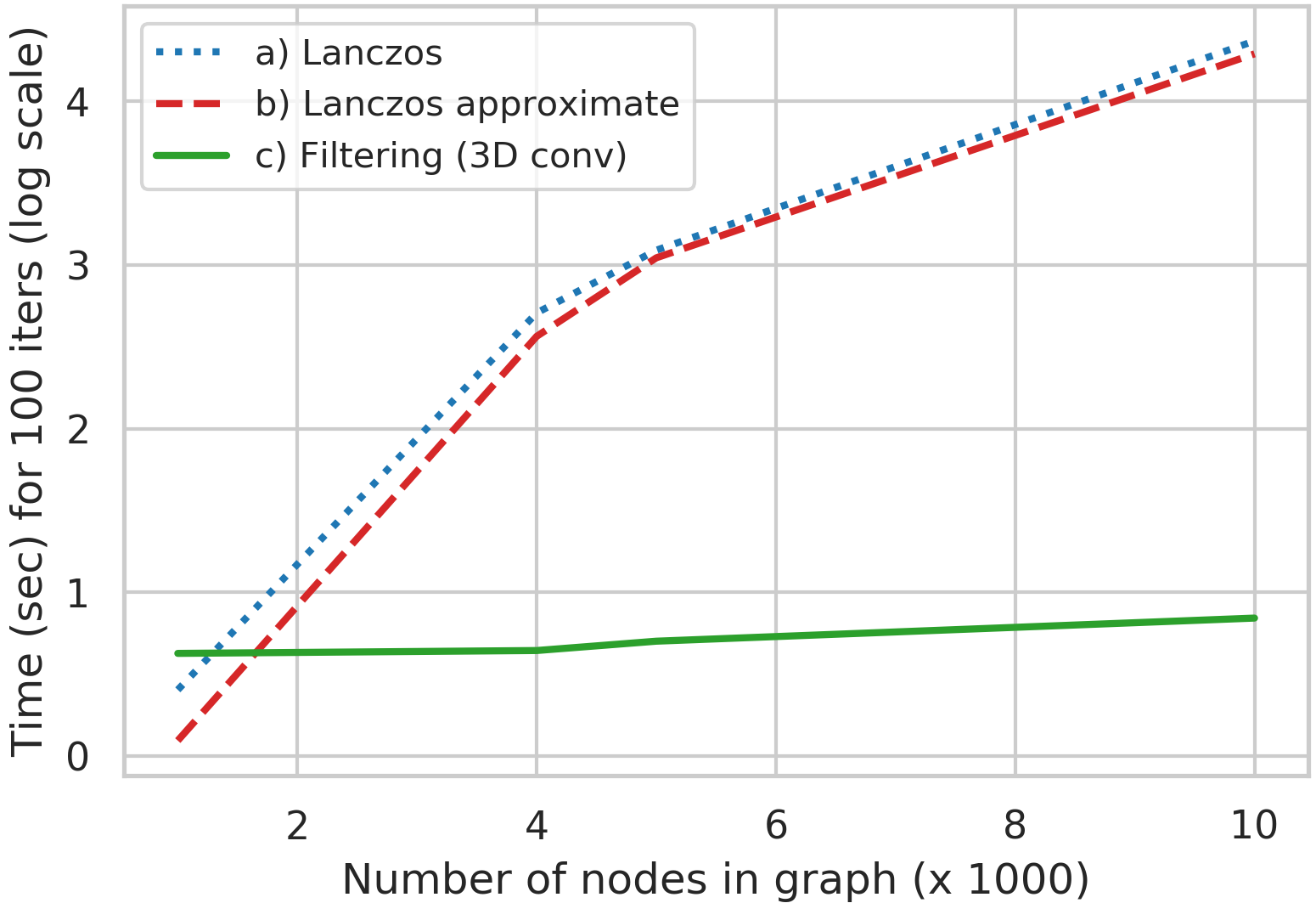}
	\end{center}
    \caption{Total runtime in logarithmic scale for 100 iterations, including the time for building the adjacency matrix for power iteration. The filtering formulation scales with the number of nodes, in contrast to power iteration, having an exponentially better time.}
    \label{fig: time_pi_fi}
\end{figure}

\begin{figure}[t]
	\begin{center}
	    \frame{\includegraphics[width=0.99\linewidth]{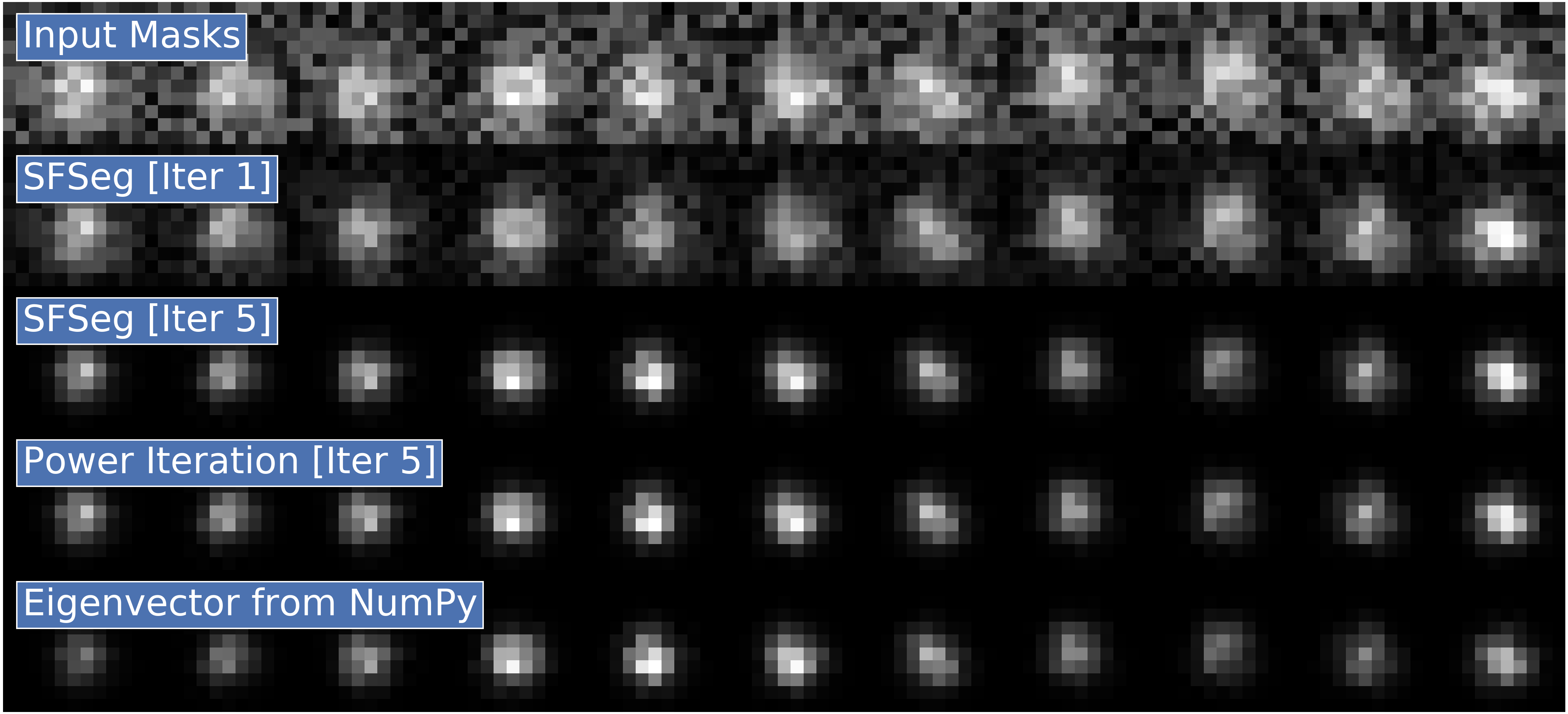}}
	\end{center}
    \caption{A toy example for qualitative comparisons with soft masks for an eleven frames video. Starting with a very noisy input segmentation mask and showing: SFSeg segmentations after 1 and 5 iterations; Power Iteration after 5 iterations; the real principal eigenvector. The results are almost identical, proving that SFSeg is a good approximation. More, for all other methods we compare with, this is tractable only on toy examples.}
    \label{fig: fi_iters}
\end{figure}

We compare the actual runtimes of three solutions: \textbf{a)} Lanczos for computing the principal eigenvector of the adjacency matrix built using the initial Eq.~\ref{eq: mij_equation} \textbf{b)} Lanczos for computing the principal eigenvector of the approximate adjacency matrix based on Eq.~\ref{eq: mij_equation_approx} \textbf{c)} our 3D convolutions approach. For a small graph of 4000 nodes (a video with 10 frames of $20 \times 20$ pixels), \textbf{a)} and \textbf{b)} have 0.15 sec/iter and \textbf{c)} our 3D filtering formulation has 0.02 sec/iter (Fig.~\ref{fig: time_pi_fi}), which is almost an order of magnitude faster. As the number of nodes grows our 3D filtering approach quickly becomes many orders of magnitude faster. The proposed method scales better and has a huge advantage when working with videos with millions of nodes. As stated before, the key reason for the large speedup is that we do not explicitly build the adjacency matrix and filtering is parallelized on GPU.

\subsection{Qualitative analysis}
We perform tests on synthetic data, in order to study the differences between the original spectral solution using the exponential pairwise scores (Eq.~\ref{eq: mij_equation}) and the one obtained after our first-order Taylor approximation trick (Eq.~\ref{eq: mij_equation_approx}). In Fig. \ref{fig: fi_iters} we see qualitative comparisons between the solutions obtained by three implementations: 1) SFSeg, 2) classical power iteration, and 3) eigenvector computed with Lanczos method in NumPy~\cite{numpy}, with original pairwise scores. The outputs are almost identical. In this synthetic experiment, the input is very noisy, but all spectral solutions manage to reconstruct the initial segmentation. 

\subsection{Quantitative analysis}
\begin{figure}[t]
\begin{center}
    \includegraphics[width=0.99\linewidth]{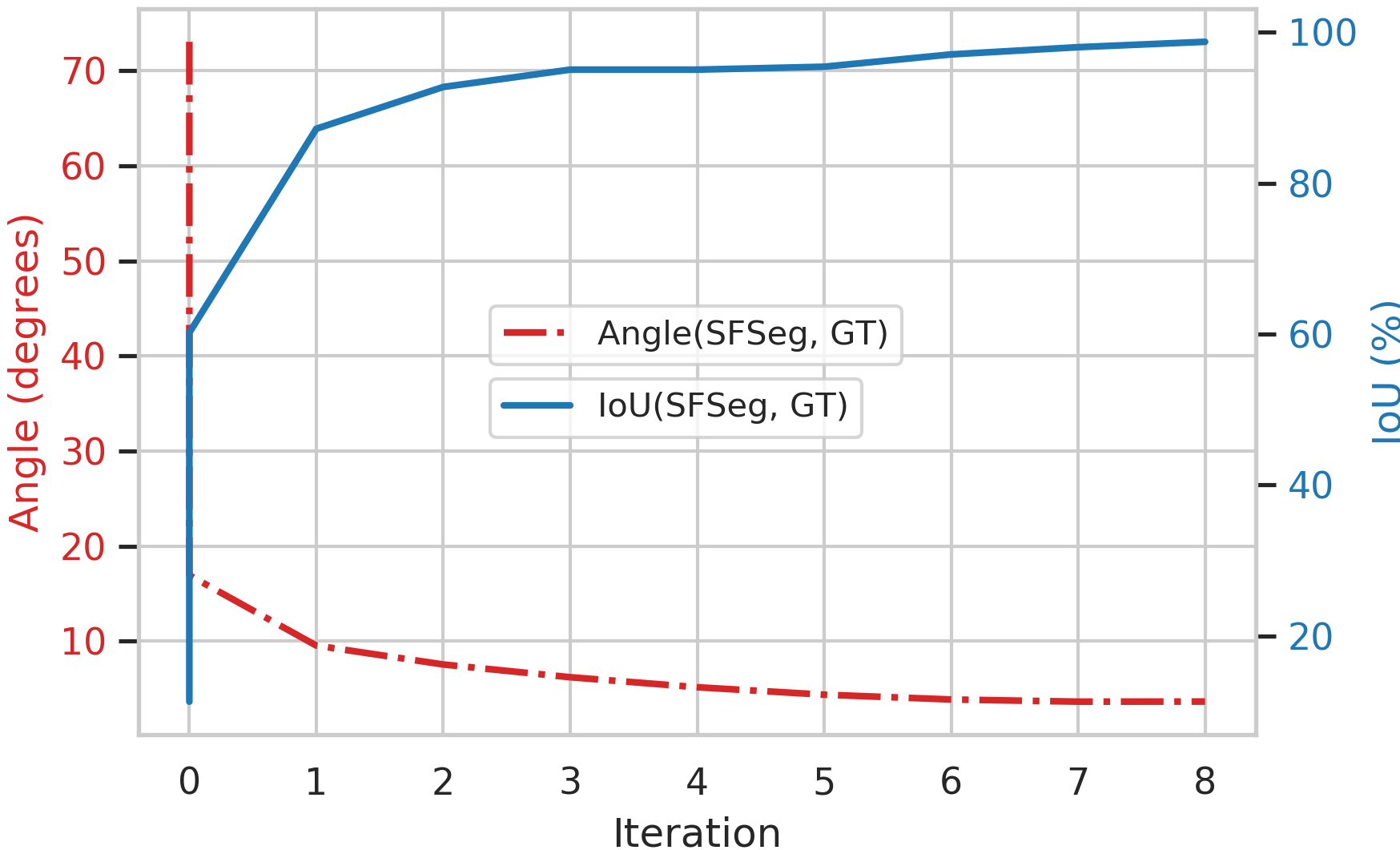}
    \end{center}
	\caption{The angle and the IoU between the exact eigenvector (computer with Lanczos method) and our SFSeg solution. The evolution of those metrics is monitored over multiple SFSeg iterations.}
	\label{fig: approx}
\end{figure}

We analyze the numerical differences between the original eigenvector and our approximation (SFSeg). We plot the angle (in degrees) and the IoU (Jaccard) between SFSeg (first-order approximation of pairwise functions, optimized with 3D convolutions) and the original eigenvector (exponential pairwise functions in the adjacency matrix), over multiple SFSeg iterations in Fig.~\ref{fig: approx}. Note that in these tests we intentionally start from a wrong solution (70 degrees difference between the SFSeg initial segmentation vector and the original eigenvector) to show that SFSeg indeed converges to practically the same eigenvector, even when the initialization is wrong.
Such comparisons can be performed only on synthetic data and relatively small videos, for which the computation of the adjacency matrix needed for the original eigenvector is tractable. The results show that SFSeg, with first-order approximations of the pairwise functions on edges and optimization based on 3D filters, reaches the same theoretical solution, while being orders of magnitude faster.

\begin{figure*}[t]
	\begin{center}
		\includegraphics[width=0.9\linewidth]{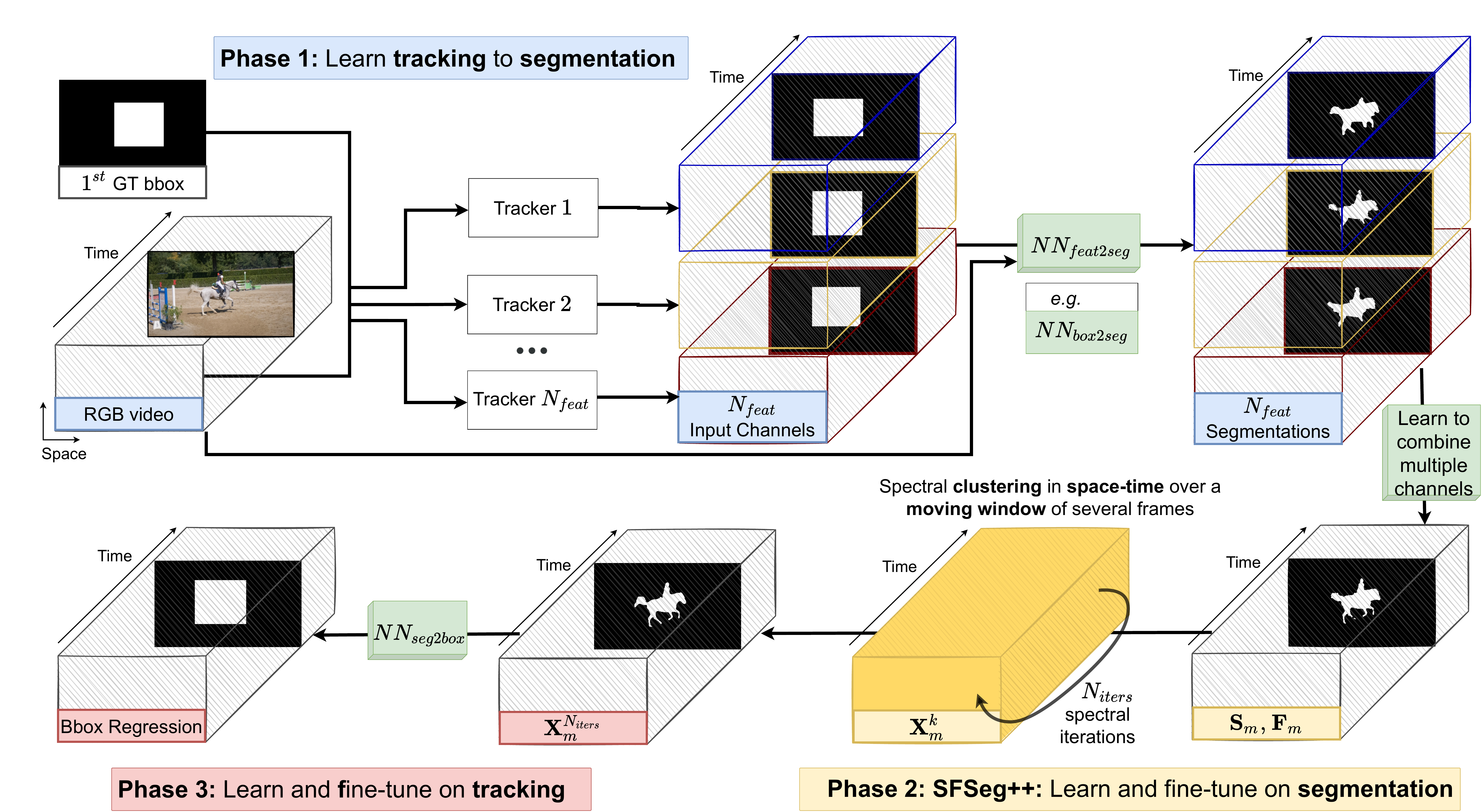}
	\end{center}
	\caption{SFTrack++: proposed object tracking system with spectral clustering in space and time.
	We start from video's RGB and $1^{st}$ frame GT bounding box of the tracked object. We run state-of-the-art trackers, in an online manner, while fine-tuning $NN_{feat2seg}$ network frame-by-frame (pretrained in Phase 1) to transform the extracted feature maps (\eg bounding boxes) to segmentation maps. Next, we learn to combine multiple segmentation inputs and refine the final mask using a spectral approach, applied also online over a moving window containing the previous N frames, for $N_{iter}$ spectral iterations (Phase 2). In Phase 3, we learn a bounding box regressor from the final segmentation mask, $NN_{seg2box}$ and fine-tune all our parameters on the tracking task. Best seen in color.}
	\label{fig: sftrack_architecture}
\end{figure*}

\subsection{Online vs Offline processing}
\label{sec:online_offline}
A full SFSeg++ iteration consists of passing through the entire video. In this form (Alg.~\ref{alg: power_iteration}), we can pass to the next iteration of the algorithm only after we go through the full video, making it an offline algorithm. That is because we need to pass (filter in 3D) through the whole video multiple times until we reach convergence.

With every iteration, local information is propagated one step further in both space and time. In practice, the method converges after several iterations. Therefore, far-away information is not that useful for determining the segmentation at a given frame. The main reason for that is that objects move and change their shape from one frame to the next, making far connections less relevant.  

\begin{figure}[t]
	\begin{center}
		\includegraphics[width=0.99\linewidth]{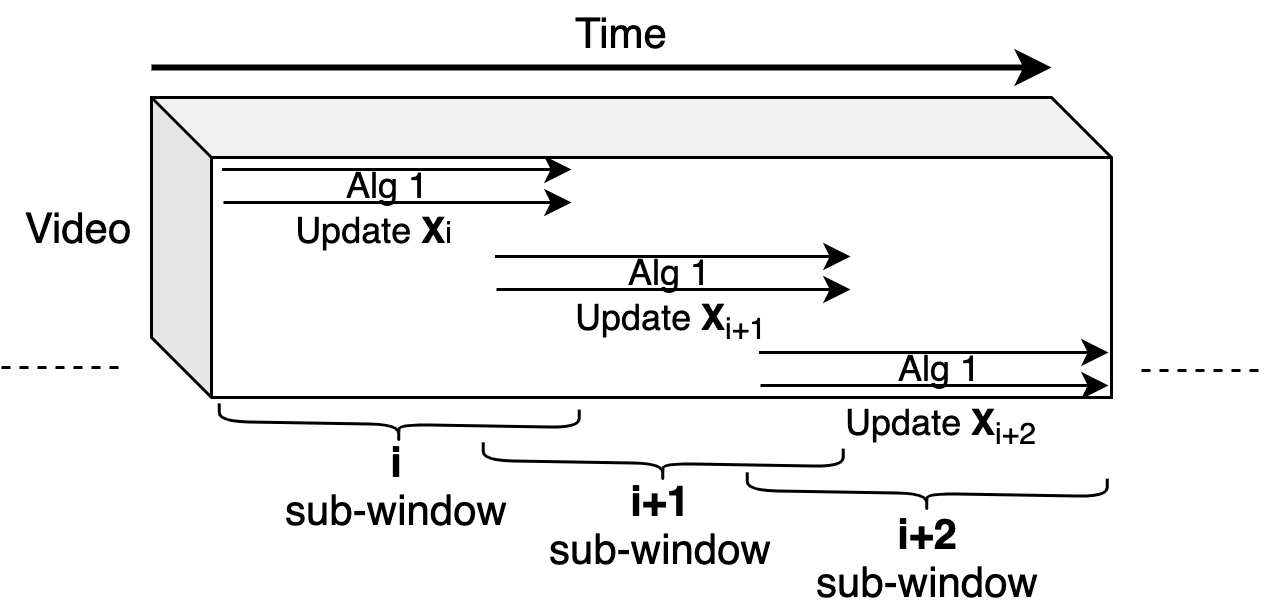}
	\end{center}
    \caption{When the video is a contiguous stream or when it is very large, instead of applying our spectral method on the full video, we can apply fewer iterations on smaller video sub-windows, with similar effect. To speed up convergence, we initialize the current solution with the final solution over the previous sub-window (for the frames that overlap).}
    \label{fig: offline_online}
\end{figure}

Based on this observation, we could expect that in practice we only need \textbf{partial iterations}, to 
apply the iterations on smaller sub-volumes of video (see Fig.~\ref{fig: offline_online}).
This also allows us to use SFSeg++ as an \textbf{online segmentation method}. We analyzed the convergence of the solution when instead of performing all iterations on the full video, we break it into a sequence with fewer iterations, in local sub-windows around a moving center frame. We observed almost identical results with the full iterations version, but at a lower computational complexity (see next paragraph). The speedup factor is identical to the size of the subwindow relative to the full video, for an equal number of iterations. Note however that all our speed comparisons to the Lanczos method are using iterations over the full video. Also, all reported results are obtained with the original full video iterations method. As stated here, the two versions, with partial and full iterations, achieve the same results for all practical purposes.

\textbf{Numerical considerations:}
the complexity of filtering in a sub-window is $\mathcal{O}(k q N_p N_i)$, where $q$ is the number of frames in a sub-window. For the online version, $q$ represents the number of frames in the past that we need to consider from the current incoming frame. It can be shown that sub-window filtering could reduce in practice the complexity of filtering over the full offline video by up to $\frac{N_f}{q}$ times. Fig.~\ref{fig: offline_online} can help visualize why this is the case.

\setlength{\tabcolsep}{0.2em}
\begin{table}[t]		
\caption{Single-channel SFSeg on DAVIS-2016 tasks. SFSeg has the same hyper-parameters per task. We also included results for other competitive (non-SOTA) inputs. We show per column: $2^{nd}$: Jaccard score for the input method; $3^{rd}$: score after denseCRF refinement $4^{th}$: score after applying SFSeg over the input method; $5^{th}$: the percentage of videos when the performance is improved after using SFSeg. The average SFSeg boost is 1.1\% in Jaccard score. \color{myblue}Using the same setting, with a common set of hyper-parameters per group, our method largely outperforms denseCRF 3D. \color{black}On average SFSeg raises performance for 80\% of videos.}
\begin{center}
	\begin{tabular}{l r c l c c}
		\toprule
         &
        \multicolumn{1}{p{2cm}}{\raggedleft Input\\Method} &
        \multicolumn{1}{p{0.8cm}}{\centering Input Score (J)} &
         \multicolumn{1}{p{1.3cm}}{\centering dense\\CRF\\(J)} &
        \multicolumn{1}{p{1.3cm}}{\centering \textbf{SFSeg}\\over\\Input (J)} &
        \multicolumn{1}{p{1cm}}{\centering Improved\\Videos\\ (\%)} \\
        \midrule
		Semi & OnAVOS~\cite{onavos}  & 86.1 & \color{myblue} 74.1 (-12.0) \color{black} & \textbf{86.3} (+0.2)& 65 \\
		Supervised  & OSVOS-S~\cite{osvoss} & 85.6 & \color{myblue} 75.1 (-10.5) \color{black}& \textbf{86.0} (+0.4)& 90\\
	    & PReMVOS~\cite{premvos} & 84.9 & \color{myblue} 79.3 (-5.6)\color{black}& \textbf{88.2} (+3.3)& 90\\
        & FAVOS~\cite{favos}  & 82.4 & \color{myblue} 73.8 (-8.6) \color{black}& \textbf{83.0} (+0.6)& 95 \\
        & OSMN~\cite{osmn}  & 73.9 & \color{myblue} 67.6 (-6.3)\color{black}& \textbf{75.9} (+2.0) & 95 \\
        \midrule
        Un & COSNet~\cite{cosnet} & 80.5 & \color{myblue} 71.3 (-9.1)\color{black}& \textbf{80.9} (+0.4) & 65 \\
		Supervised  & MotAdapt~\cite{motadapt} & 77.2 & \color{myblue} 67.3 (-9.9) \color{black}& \textbf{77.5} (+0.3) & 65 \\
		& PDB~\cite{pdb}  & 77.2 & \color{myblue} 68.9 (-8.3) \color{black}& \textbf{77.4} (+0.2) & 60\\
		& ARP~\cite{arp} & 76.2 & \color{myblue} 68.2 (-8.0) \color{black}& \textbf{77.7} (+1.5) & 90 \\
	    & LVO~\cite{lvo}  & 75.9 & \color{myblue} 68.7 (-7.2)\color{black}& \textbf{78.8} (+2.9) & 90 \\
	    & FSEG~\cite{fseg} & 70.7 & \color{myblue} 65.2 (-5.4)  \color{black}& \textbf{72.3} (+1.6) & 95\\
	    & NLC~\cite{nlc}  & 55.1 & \color{myblue} 50.0 (-5.1) \color{black}& \textbf{55.6} (+0.5) & 65\\
		\midrule
		& Average Boost  &  & & +1.1\% & 80\%\\
		\bottomrule
    \end{tabular}
\end{center}	
\label{tab: davis_results}
\end{table}

\subsection{Spectral Segmentation integrated in Tracking}
We now take a step beyond the segmentation task and propose SFTrack++, an object tracking solution that integrates our previous spectral segmentation in space and time, into tracking. Our main motivation is the current need in object tracking for more refined object shapes, beyond the simplistic and often inaccurate object bounding boxes. To better capture complex aspects of the tracked object, we learn over the multi-channel formulation, which permits different points of view for the same input. The channel inputs are in our experiments (Sec.~\ref{subsec: exp_sftrack++}), the output of multiple tracking methods. After combining them, instead of relying only on hidden layers representations to predict a good tracking bounding box, we explicitly learn an intermediate, more refined one, namely the segmentation map of the tracked object. This prevents the rough common bounding box approach to introduce noise and distractors in the learning process.

Our tracking algorithm has three phases, as we visually present them in Fig.~\ref{fig: sftrack_architecture}. In \textbf{Phase~1}, we learn a neural net, $NN_{feat2seg}$, that transforms the RGB and a frame-level feature map extracted using a tracker (\eg bounding box from a tracker prediction) into a segmentation mask. Using only the RGB as input is not enough, because frames can contain multiple objects and instances, and we also need a pointer to the tracked object to predict its segmentation. Next, in \textbf{Phase~2}, we run multiple state-of-the-art trackers frame-by-frame over the input as an online process and extract input channels from them (\eg bounding boxes). We transform those feature maps to segmentation maps with the previously recalled module, $NN_{feat2seg}$. Next, we learn to combine and refine the outputs for the current frame using a spectral solution for preserving space-time consistency, adapted to learn over multiple channels. Note that, when applying the spectral iterations, we use a sliding window approach over the previous N frames in the video volume. For supervising this path, we use segmentation ground-truth. \textbf{Phase~3} learns a neural net as a bounding box regressor over the final segmentation map from the previous phase, $NN_{seg2box}$, while fine-tuning all the other trainable parameters in the model, using tracking GT.

\section{Experimental Analysis}
\label{sec: experiments}

\subsection{Spectral Segmentation}

\subsubsection{\textbf{Single-channel SFSeg on DAVIS-2016}} DAVIS-2016~\cite{davis2016} is a densely annotated video object segmentation dataset. It contains 50 high-resolution video sequences (30 train/20 valid), with a total of 3455 annotated frames of real-world scenes. The benchmark comes with two tasks: the unsupervised one, where the solutions do not have access to the first frame of the video, and the semi-supervised one, where the methods use the ground-truth from the first frame.

We test the single-channel version, SFSeg, with input from pre-computed segmentations of the video produced by top methods from DAVIS-2016, on both tasks. For the features maps, we initialized $\mathbf{S}$ with the pre-computed input segmentation values. For $\mathbf{F}$, we used two channels: the magnitude for the direct optical flow and for the reverse optical flow. For optical flow we used \cite{flownet2-pytorch} implementation of Flownet2~\cite{flownet2}. We set: $N_{i} = 5$; $\alpha = 1$ and $p = 0.1$ for unsupervised task and $p = 0.2$ for the semi-supervised one. The algorithm is implemented as in Alg.~\ref{alg: power_iteration}. For the temporal consistency metric, we used the same optical flow implementation.

\settasks{
  label-offset = 0em ,
  item-indent = 0em ,
  item-indent = 1em ,
  column-sep = 1em
}

\begin{figure}[t]
	\begin{center}
		\includegraphics[width=0.99\linewidth]{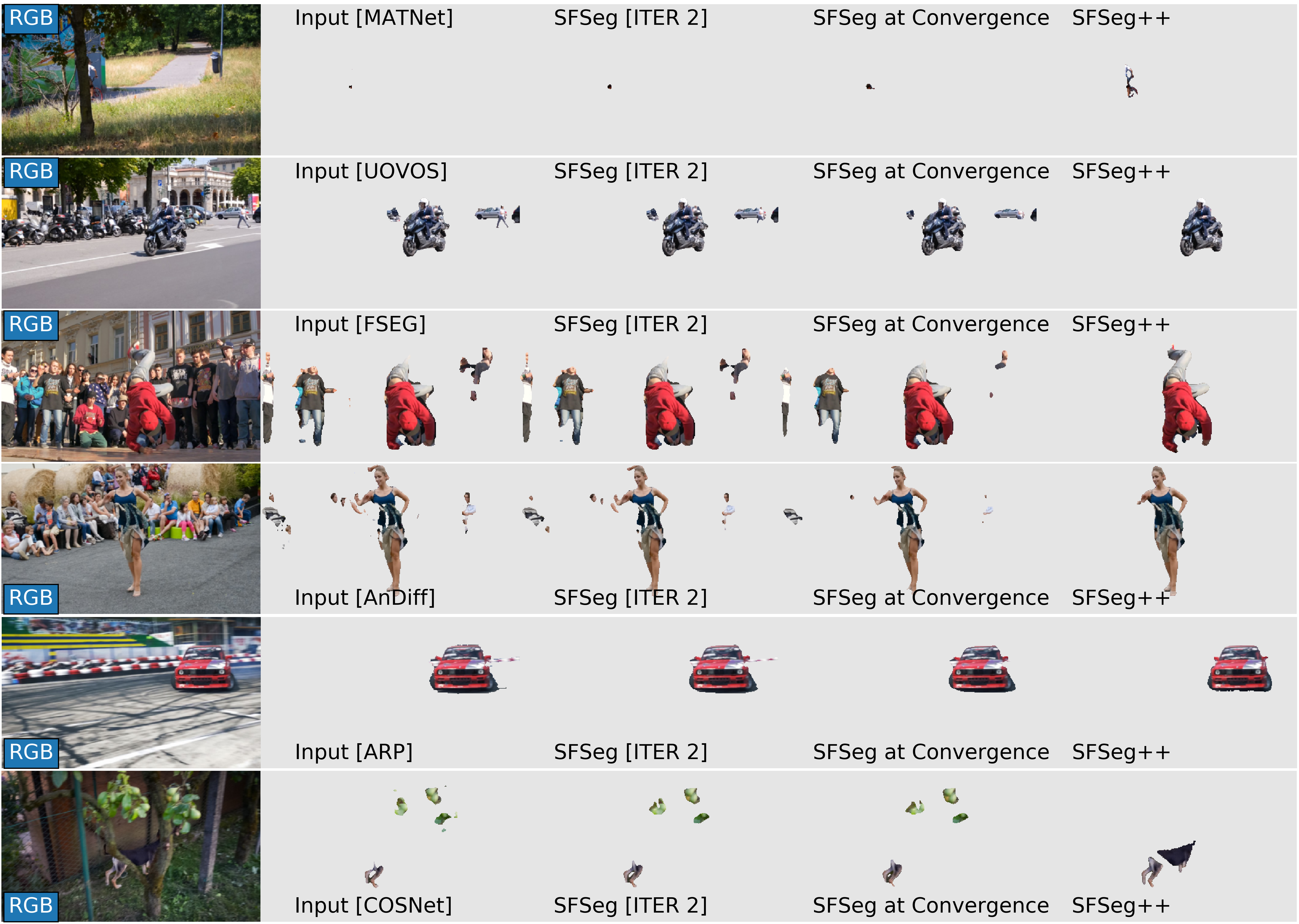}
	\end{center}
	\caption{We present the evolution of the single-channel SFSeg towards convergence, over several iterations, starting from the output of different methods, as stated in the square parenthesis. Using the input segmentation mask (column 2) from top methods: \cite{matnet, uovos, fseg, andiff, arp, cosnet}, we show the intermediate value of the mask at the second iteration (column 3) and on convergence (iteration 5, column 4). In the last column, we also show for comparison, the output of the learned multi-channel SFSeg++ (with 15 input channels, from 15 different methods). Note the vastly superior results of SFSeg++, which learns to combine many channels, learning both to recover occluded objects (1st and last line), but also to remove distractors (lines 2-5). Best seen in color.}
	\label{fig: qualitative_iterations}
\end{figure}

In Tab.~\ref{tab: davis_results} we show the results of our method, SFSeg, when combined with top methods on DAVIS-2016, semi-supervised and unsupervised tasks. For a better understanding of the results, we also show the effect of applying SFSeg over other competitive, non-SOTA methods. Note that the improvement is not related to the quality measure of the input: in some cases, the improvement is stronger when input comes from stronger methods. Nevertheless, we consistently improve over the input method, whose segmentation mask we use to initialize the segmentation $\mathbf{X}_0$ (Eq.~\ref{eq: matrix_full_eigenvector}). \color{myblue}We show in parallel the results for denseCRF~\cite{denseCRF}, a widely used refinement method, known to perform well when strongly fine-tuned. For a fair comparison, we kept the same hyper-parameters for all methods, using their official implementation (Gaussian kernel of $3\times3\times3$). The results show a catastrophic drop in denseCRF performance. The average time for running denseCRF is $80$ times larger when compared with SFSeg ($0.8$ vs $0.01$ sec. per frame). \color{black}

\begin{figure*}[!t]
\begin{center}
    \frame{\includegraphics[width=0.9\textwidth]{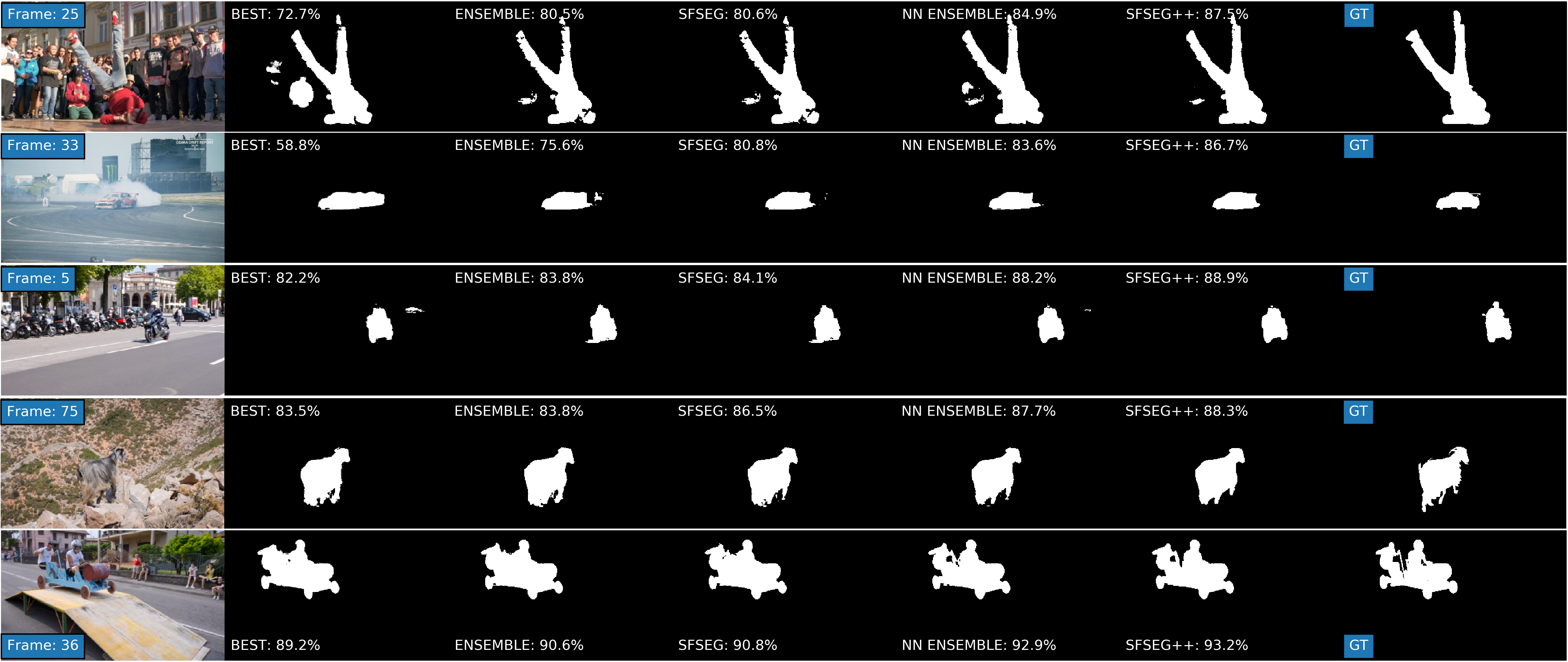}}
	\caption{Qualitative output for several videos in DAVIS-2016, from left to right: the best method
	(which is also part of the ensemble), the ensemble output (arithmetic mean over 15 top methods), SFSeg applied over the ensemble output, the learned ensemble output with neural nets and finally, our SFSeg++. On the last column we also show the ground-truth (GT) for comparison. Both SFSeg and SFSeg++ improve significantly (especially the multi-channel learnable SFSeg++) over their input channels. Best seen in color.}
	\label{fig: qualitative_hard}
\end{center}
\end{figure*}
In Fig.~\ref{fig: qualitative_iterations} we show the iterative effect of SFSeg. Each example starts from the initial RGB frame and its initial segmentation (as produced by top DAVIS-2016 methods), and presents the segmentation at an intermediate iteration and the final one, when SFSeg reaches convergence. For comparison, we also show the output of the learnable multi-channel SFSeg++. Note how SFSeg++ learns to get the best of the combination of multiple inputs, each with significantly lower performance. See how the iterative 3D spectral filtering process of SFSeg improves from one iteration to the next.

In Fig.~\ref{fig: qualitative_hard} we show comparative qualitative examples between the output of the best method in the ensemble (MATNet~\cite{matnet}), the arithmetic mean, SFSeg over this mean, the learned ensemble, and SFSeg++.\\

\subsubsection{\textbf{Single-channel SFSeg on SegTrackv2}}
\label{subsec: segtrack}
SegTrackv2~\cite{segtrack2} is a video object segmentation dataset, containing 14 videos, with multiple objects per frame. The purpose for video object segmentation task is to find the segmentation for all the objects in the frame, either by using the first frame or not. We use our standalone method, SFSeg, applied over the soft output of a competitive Backbone (BB): UNet~\cite{unet} over ResNet34~\cite{resnet} pretrained features, fine-tuned 40 epochs on salient object segmentation in images on DUTS dataset~\cite{saliencyDS}, using RectifiedAdam as optimizer. In Tab.~\ref{tab:segtrack2} we show comparative results of our standalone method and other top solutions on the SegTrackv2 dataset.\\

\begin{table}[t]
\caption{Comparative results on SegTrackv2. Our standalone solution, BB (Backbone) + SFSeg + denseCRF, obtains the best results among the other top methods in the literature.}
\begin{center}
	\begin{tabular}{r r r r}
		\toprule
		\multicolumn{1}{r}{Method} &
        \multicolumn{1}{r}{Score (J)}\\
        \midrule
        LVO~\cite{lvo} & 57.3\\
        FSEG~\cite{fseg} & 61.4\\
		OSVOS~\cite{osvos} & 65.4\\
		NLC~\cite{nlc}  & 67.2\\
		MaskTrack~\cite{masktrack} & 70.3\\
		Ours: BB + SFSeg + denseCRF & \textbf{72.7} \\
        \bottomrule
    \end{tabular}
\end{center}			
\label{tab:segtrack2}
\end{table}

\subsubsection{\textbf{Standalone Single-channel  SFSeg vs denseCRF}} We compare SFSeg with denseCRF~\cite{denseCRF}, which is one of the most used refinement methods in video object segmentation \cite{pdb}. When applied over the same backbone presented above, we observe that SFSeg brings a stronger improvement than denseCRF on both DAVIS-2016 and SegTrackv2 (Tab.~\ref{tab:backbone}). Moreover, the two are complementary: in combination, the performance is boosted by the largest margin.\\

\begin{table}[t]
\caption{Refinement Comparison \color{myblue} on Standalone BackBone \color{black}. We compare SFSeg with denseCRF when applied to a competitive end-to-end Backbone (BB), as detailed in Sec.~\ref{subsec: segtrack}. While SFSeg outperforms denseCRF when used individually, the two methods prove to be not only different, but also complementary, since combining them boosts the Jaccard score.}
\begin{center}
	\begin{tabular}{l r r r}
		\toprule
		\multicolumn{1}{l}{Method} &
		\multicolumn{1}{p{1cm}}{\raggedleft DAVIS (J)} &
		\multicolumn{1}{p{2.1cm}}{\raggedleft SegTrackv2 (J)} \\
        \midrule
		BB & 67.2 & 72\\
		BB + denseCRF & 68.1 & 72\\
		BB + SFSeg& 68.7 & 72.1 \\
		BB + SFSeg + denseCRF & \textbf{69.2} & \textbf{72.7} \\
		\bottomrule
    \end{tabular}
\end{center}
\label{tab:backbone}
\end{table}

\subsubsection{\textbf{Learned Multi-channel SFSeg++: Experimental Setup}}
\label{subsec: sfseg++-experimental-setup}

Before we present the actual tests performed with SFSeg++, we provide more details about the exact experimental setup. In the comparative experiments, SFSeg++ uses, as input feature channels, the outputs of the top $15$ methods, as published on DAVIS-2016 website, for the task where no ground-truth is given in the first frame of the test videos (officially labeled the "Unsupervised task"): MATNet~\cite{matnet}, AnDiff~\cite{andiff}, COSNet~\cite{cosnet}, ARP~\cite{arp}, UOVOS~\cite{uovos}, FSeg~\cite{fseg}, LMP~\cite{lmp}, TIS~\cite{tis}, ELM~\cite{elm}, FST~\cite{fst}, CUT~\cite{cut}, NLC~\cite{nlc}, MSG~\cite{msg}, KEY~\cite{key}, CVOS~\cite{cvos}, TRC~\cite{trc}. The experiments which test the ability of our learning approach to deal with noisy input channels, modify these initial channels according to the different types of noise used, as explained in the next Section ~\ref{subsec: sfseg++-noisy-input}. Also, in all experiments, we train all models in a supervised fashion only on the DAVIS-2016 training split. Results are, of course, reported on DAVIS-2016 validation set. We follow the exact official protocol of DAVIS-2016 challenge. For efficiency and compactness, we noted that in the case of SFSeg++ we can use the same $15$ initial channels to construct both unary and pairwise maps. Thus, $\mathbf{S}_i = \mathbf{F}_i$ and their learned weights are also shared: $w_{s, i} = w_{f, i}$. \\

\subsubsection{\textbf{Applying Multi-channel SFSeg++ over Noisy Input}}
\label{subsec: sfseg++-noisy-input}
To better understand the effectiveness of learning over multiple input channels we first analyze our method's robustness to noise. Thus, we added different types of noise to the input channels.
First, we keep only the top 3 original input methods, replacing the rest 12 methods with noise. Secondly, we only disturb the rest 12 methods with noise. Thirdly, we used a combination of the two: 5 methods disturbed and 7 replaced with noise.
We tested with three types of noise: \textbf{a) Uniform (U)} - This introduces a low to moderate level of noise. We add uniform noise in $[{0, 0.1}]$, uniformly over the whole input map (which has values between $0$ and $1$ and normalize the result, \textbf{b) Salt\&Pepper (SP)} - This noise is more invasive for the original maps. We randomly replace pixels with $0$ or $1$, with 20$\%$ probability and \textbf{c) Black and White randomly placed Rectangles (BWR)} - This is an aggressive and more structured type of noise, to simulate the effect of an occlusion or per frame distractor. We add one rectangle per frame, uniformly sampling its position (inside 100px margins) of randomly sized weight and height, filling between 2\% to 5\% of the total frame size. 
We present the results and comparisons in Tab.~\ref{tab: noisy_exp}, where for the basic Ensemble we use the average output of the $15$ different methods and for the Best Learned Ensemble we selected among the different Deep Neural Nets architectures that learn over the same 15 input channels, over a certain temporal window (see Sec.~\ref{subsec: comparison_NN_ensembles} for more details).

\begin{figure}[t]
	\begin{center}
	    \includegraphics[width=0.99\linewidth]{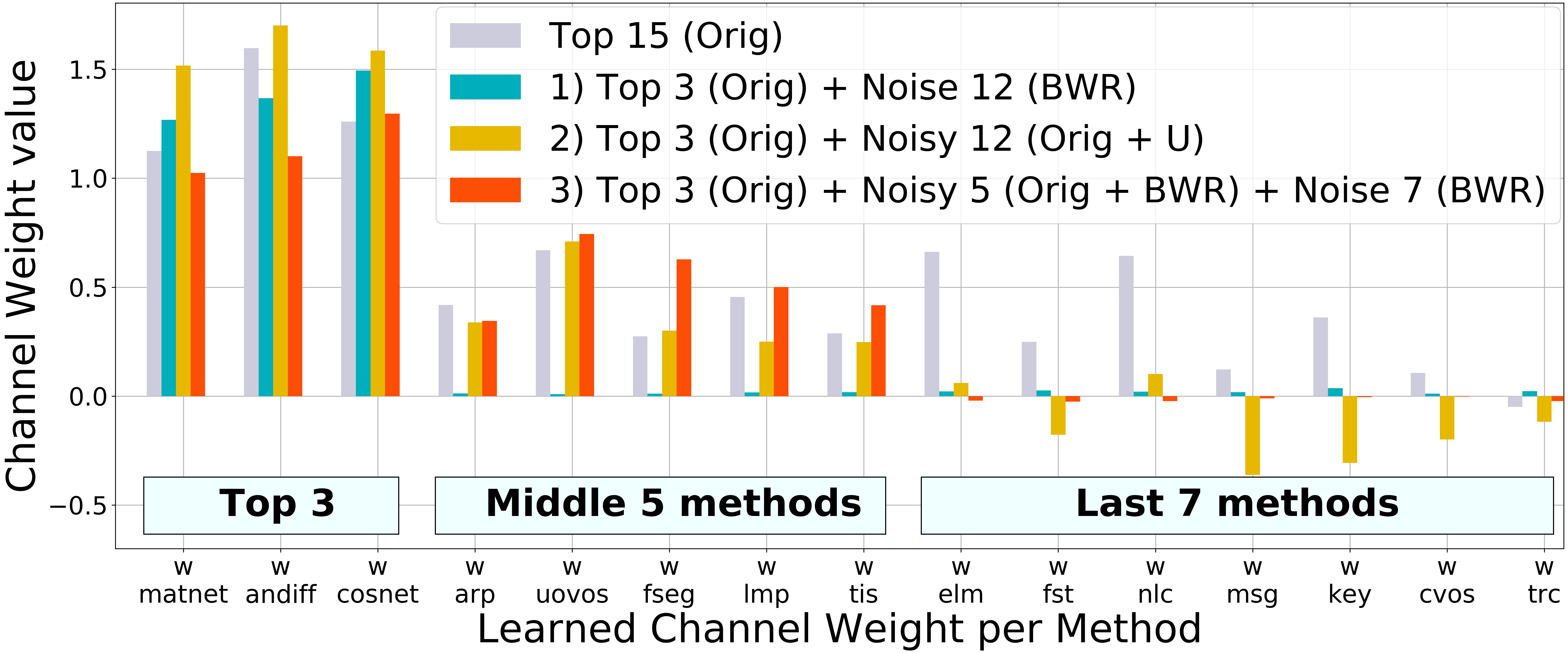}
	\end{center}
    \caption{Learned Channels Weights: We display the learned weights values for SFSeg++ learning over unmodified input channels from top 15 methods as well as the weights for the different cases of added noise (as explained in Sec.~\ref{subsec: sfseg++-noisy-input}). The input methods are shown from left to right, in descending order of their individual performance. We immediately notice that the learned weights are larger for channels from methods with superior performance (towards the left: \textbf{top 3 vs the remaining 12} or \textbf{middle 5 vs last 7}) or channels not corrupted by noise (\textbf{top 3 vs the remaining 12}). This indicates that learning is meaningful: it is an efficient and automatic way to select the relevant input channels. Best seen in color.}
   \label{fig: learned_weights}
\end{figure}

In Fig.~\ref{fig: learned_weights} we show the actual weights learned for the different cases of added noise or for the case of using the original input channels. Note that the input methods (channels) are presented in the descending order of their individual performance. We can immediately notice that SFSeg++ learns to select the most relevant channels (larger weights) and ignore the more noisy or weaker ones (lower weights). For example, when no noise is added (Fig.~\ref{fig: learned_weights} - Top 15), the weights are correlated with performance, thus channels with better performance (on the left side of the plot) tend to have larger weights than the ones with weaker performance (towards the right side of the plot). When noise is added to some channels (labeled as "Noisy" in Fig.~\ref{fig: learned_weights}) or some channels are completely replaced with noise (labeled as "Noise" in Fig.~\ref{fig: learned_weights}), the top 3 methods (without noise) receive very strong weights, while the remaining ones much weaker weights. It is also interesting that for the same type and amount of noise being added, the middle 5 channels have larger weights than the last 7 channels, as they have better individual performance (case 2). Also, when noise is added to some channels while others are replaced by noise (case 3), the noisy ones receive significantly stronger weights than the ones that are complete noise, with close-to-zero weights. The visual analysis of the weights reveals that learning is indeed meaningful: it automatically discovers which channels are relevant and which ones are not and weighs them accordingly.

\noindent \textbf{Key observations:} When interpreting the learned weights of the input channels, we note that the channels having noise added receive very low weights and become mostly ignored. The selection of channels taking place during learning focuses more on top methods or channels without noise. This shows the robustness of our learning. We also note that while the single-channel SFSeg is able to improve significantly over poor inputs (SP), SFSeg++ truly distances itself from even the best-learned ensembles. SFSeg++ achieves top performance in all cases. Thus, SFSeg++ learns to exploit relevant details in input maps better than any competitor, due to the combination of learning and clustering over the time dimension.\\

\begingroup
\setlength{\tabcolsep}{2pt} 
\begin{table}
\caption{SFSeg++ Noisy Input: The experiments are done for 3 types of noise: Uniform (U), Salt\&Pepper (SP), Black and White Rectangles (BWR) (see Sec.~\ref{subsec: sfseg++-noisy-input} for details). We test the performance of SFSeg++ in 3 combinations of the noise with the 15 input methods. Note that the largest difference between SFSeg++ and the best learned NN ensemble is on Uniform noise cases, where the input is less affected by noise. This proves that training with SFSeg++ over several input channels exploits more relevant details in input compared with other learned ensembles. It also shows the effectiveness of clustering over the temporal dimension.}
\begin{center}
	\begin{tabular}{l l r r r r}
		\toprule
		\multicolumn{1}{p{0.2cm}}{Noise Type} &
		\multicolumn{1}{p{2.cm}}{Noise Combination} &
		\multicolumn{1}{p{0.9cm}}{\raggedleft Ensemble} &
		\multicolumn{1}{p{1.2cm}}{\raggedleft SFSeg over Ensemble} &
		\multicolumn{1}{p{1.2cm}}{\raggedleft Best Learned Ensemble} &
		\multicolumn{1}{p{1cm}}{\raggedleft SFSeg++} \\
        \midrule
        &Top 3 + Noise 12             & 80.6 & 80.7 & 84.8 & \textbf{85.1} \\
        \textbf{U} &Top 3 + Noisy 12             & 83.5 & 83.7 & 84.4 & \textbf{85.4} \\
        &Top 3 + Noisy 5 + Noise 7    & 84.2 & 84.2 & 84.9 & \textbf{85.2} \\
        \midrule
        &Top 3 + Noise 12            & 61.5 & 62.5 & \textbf{85.5}
        & \textbf{85.5} \\
        \textbf{SP} & Top 3 + Noisy 12            & 81.0 & 83.3 & \textbf{84.9} & \textbf{84.9} \\
        &Top 3 + Noisy 5 + Noise 7  & 83.8 & 84.0 & \textbf{85.0} & \textbf{85.0} \\
        \midrule
        
        &Top 3 + Noise 12          &  79.6 & 80.2 &  85.5 & \textbf{85.6} \\
        \textbf{BWR} &Top 3 + Noisy 12            & 82.2 &  83.0 &  84.8 & \textbf{84.9}\\
         &Top 3 + Noisy 5 + Noise 7  & 84.3 &  84.3 & \textbf{85.9} & \textbf{85.9}\\
        
        \midrule
        \textbf{None} & Top 15    & 83.5 & 83.6 & 85.5 & \textbf{86.6} \\
		\bottomrule
    \end{tabular}
\end{center}
\label{tab: noisy_exp}
\end{table}
\endgroup

\subsubsection{\textbf{Comparing SFSeg++ with learned neural net ensembles}}
\label{subsec: comparison_NN_ensembles}

To better understand the value of the space-time clustering approach combined with learning as opposed to simply learning to combine multiple input channels, we compared against some powerful deep neural networks ensembles, which take as input the same channels as SFSeg++. We varied the number of parameters and architectures of the deep networks, to make sure the comparisons do not depend on the specific neural net size and architecture.

The learning task was for the net to learn to predict the final segmentation, in a supervised way, using the DAVIS-2016 training set for which ground-truth segmentations are available. Therefore, SFSeg++ and all neural net ensembles had exactly the same input (at testing and training) and the same supervised training set. Also note that the neural net ensembles also had access to inputs over several time frames (5-frames temporal windows), thus the temporal dimension was not absent from the neural net ensembles.  Of course that 
SFSeg++ has access to the full video, indirectly through the space-time power iteration process, but, as stated previously (Sec \ref{sec:online_offline}) it is not the width of the temporal window that matters most, but rather the iterative clustering procedure. Moreover, during the SFSeg++ training phase, for efficiency and without loss in performance, we only consider sub-videos of 5 consecutive frames, which are in fact identical to the ones considered by the most powerful learned ensembles.

We started creating simple and then more complex neural net ensembles, w.r.t architecture, and the number of parameters. Interestingly enough, we found out that a deep net ensemble with many parameters is not so strong as one with fewer parameters. Perhaps this is not that surprising considering how small the training set is (30 training videos), so large nets are prone to overfitting on this small training set. Moreover, note that the arithmetic mean is usually a very strong baseline when we average over multiple top methods.

For choosing a good neural net ensemble for segmentation, we started from UNet~\cite{unet} architecture for 2D (per frame) and adapted it for 3D (per 5 frames), with a various number of parameters ($380$ - $1.4$ millions). For the small network ensembles we stacked $2$ to $3$ conv layers (with kernel size $1, 3, 5$) followed by ReLU non-linearity. As in the case of SFSeg++, we optimize the Focal-Dice loss~\cite{focal} (with $0.75$ coefficient) with an AdamW optimizer with AMSGrad~\cite{adamw-amsgrad} from Pytorch~\cite{pytorch} and a ReduceLROnPlateau scheduler (with a reduce factor of $0.5$) and a strong L2 regularization. For each architecture we choose the best: learning rate from (1e-1, 1e-2, 1e-3, 1e-4, 1e-5), L2 regularizer from (5e-1, 1e-2, 5e-2, 1e-3, 5e-3) and scheduler patience for reducing the learning rate, from ($5, 10$ epochs).

Tab.~\ref{tab: ensemble_choice} confirms that the 3D versions perform significantly better compared with the 2D ones. As mentioned above, the experiments also come with the interesting result that under our task assumptions (learn an ensemble over very strong state-of-the-art input methods with a small training and validation sets), the smaller nets often perform better.

\noindent \textbf{Key observations:}
The comparisons between the learnable multi-channel SFSeg++ and different types and sizes of neural net ensembles show a clear advantage for SFSeg++, which is due to several factors: 1) the ability of SFSeg++, through space-time clustering, to better take advantage of the temporal dimension, naturally and with minimal training data required. 2) the ability of SFSeg++ to not overfit for small training sets, given its small set of learnable weights as compared to the much larger neural nets. Note, however, as also stated in the theoretical section on learning (Sec. \ref{subsec: learning}), that in principle SFSeg++ could also learn its input deep feature extractors, end-to-end, if needed, as long as they are learnable (differentiable: \eg deep nets) themselves.\\

\begin{table}
\caption{Comparison to learned neural net ensembles: we trained various ensembles' net architectures (2D, 3D, UNet based, shallow NN), with different numbers or parameters, using a relatively small training set of 30 videos, starting with the same input maps from the top 15 input methods as SFSeg++. Note that SFSeg++ outperforms the best learned NN ensembles by 1.1\%, with only 16 learned parameters (one parameter per input channel and the bias term).}
\begin{center}
	\begin{tabular}{l r r}
		\toprule
		\multicolumn{1}{l}{NN Ensemble Method} &
		\multicolumn{1}{p{2cm}}{\raggedleft Number of parameters} &
		\multicolumn{1}{p{2cm}}{\raggedleft Jaccard (\%)} \\
        \midrule
		Unet3D large & 1.4 mil & 85.4 \\
		Unet3D small & 330k & 85.4 \\
		Unet2D & 140k & 83.9\\  
		Shallow2 2D & 35k & 84.8 \\
		Shallow3 2D & 5k & 84.0 \\
		Shallow4 2D & 380 & 85.5\\
		\textbf{SFSeg++} & 16 & \textbf{86.6}\\
		\bottomrule
    \end{tabular}
\end{center}
\label{tab: ensemble_choice}
\end{table}

\subsubsection{\textbf{Learned Multi-channel SFSeg++ on DAVIS-2016}}
\begin{figure}[t]
	\begin{center}
	    \includegraphics[width=0.99\linewidth]{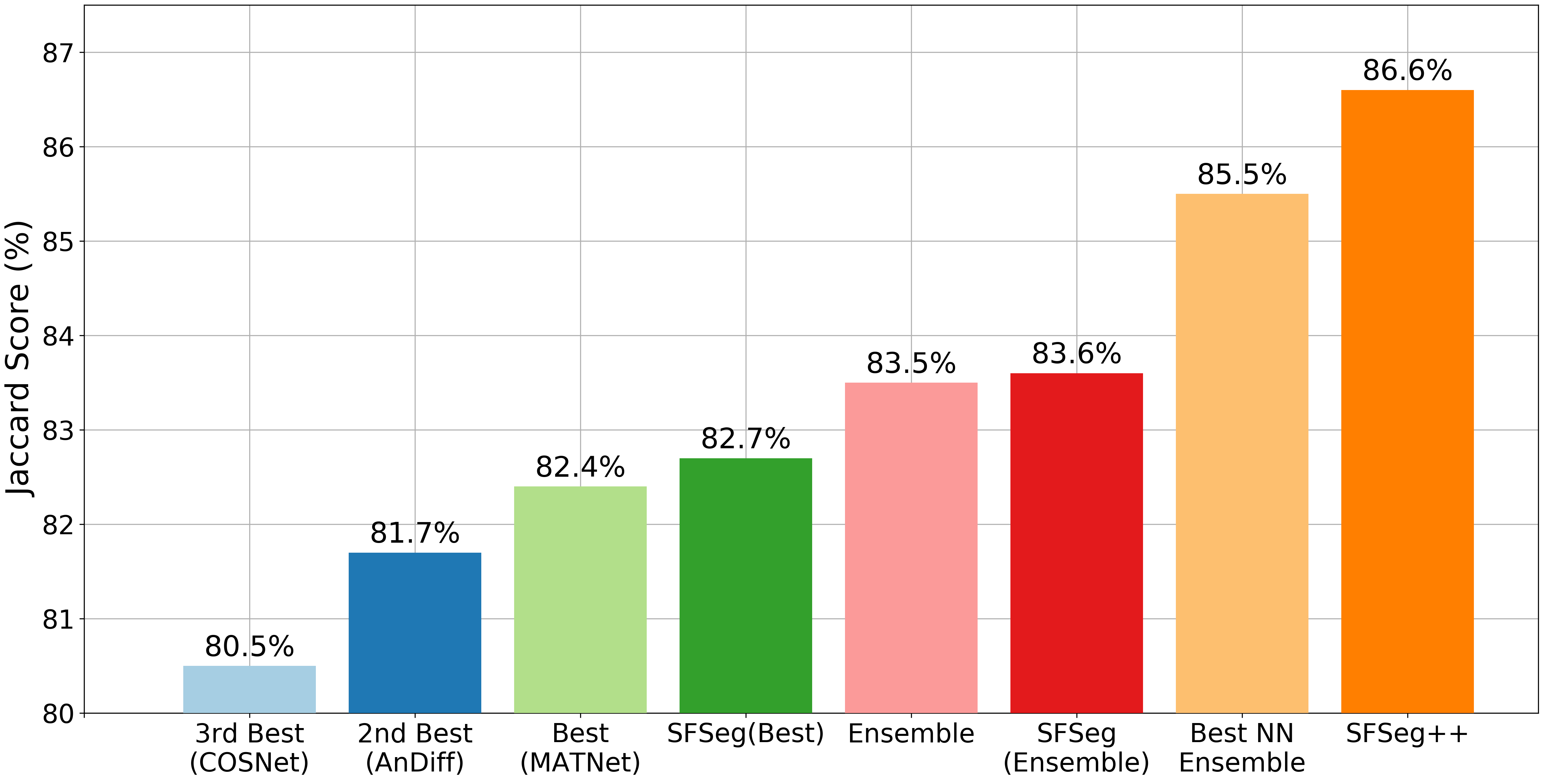}
	\end{center}
    \caption{Experimental comparisons with plot bars: to better understand the difference in performance between various methods, we show them here, as plot bars, in ascending order, starting from the top 3 single methods, followed by SFSeg on the Best method, a basic average Ensemble, SFSeg applied over the average Ensemble, the Best learned neural net (NN) Ensemble and ending with SFSeg++. While SFSeg improves over single-channel input methods, SFSeg++ boost is significantly larger, also when compared with the Best learned NN Ensemble. Best seen in color.}
    \label{fig: order_methods}
\end{figure}

This experiment shows the improvements SFSeg++ brings over top approaches: a learned NN ensemble (as presented in the previous section), basic ensemble (average of 15 single channel outputs)
and top single-channel methods. All training was performed only over the DAVIS-2016 trainset of 30 videos, with ground-truth segmentations for all frames available. Again, all methods
receive the same input from 15 segmentation methods, as previously described in Sec.~\ref{subsec: sfseg++-experimental-setup}. We report results in Tab.~\ref{tab: main_learning_exp} and Fig.~\ref{fig: order_methods}.

\noindent \textbf{Key observations:} While SFSeg outperforms the current single state-of-the-art in DAVIS-2016, the learnable multi-channel SFSeg++ outperforms the best learnable NN ensembles by a solid $3.1\%$. It is interesting to note that SFSeg alone, without learning applied over the single-channel output of the standard average ensemble, further improves over the ensemble. Then, SFSeg++, which has the ability to learn over multiple input channels, outperforms the best learned NN ensemble. These two results strongly indicate the complementary value that learning and iterative space-time clustering bring, over the more traditional feed-forward pass in the conv neural nets, which do not take full advantage of the temporal dimension. These experiments along with the ones from the previous Sec.~\ref{subsec: comparison_NN_ensembles} also show that in the case of small supervised training data, a general, non-specific, data-independent space-time clustering approach combined with a small set of learned weights could be significantly more effective than models that rely heavily on supervised training data.\\

\begin{figure*}[t]
	\begin{center}
        \includegraphics[width=0.9\linewidth]{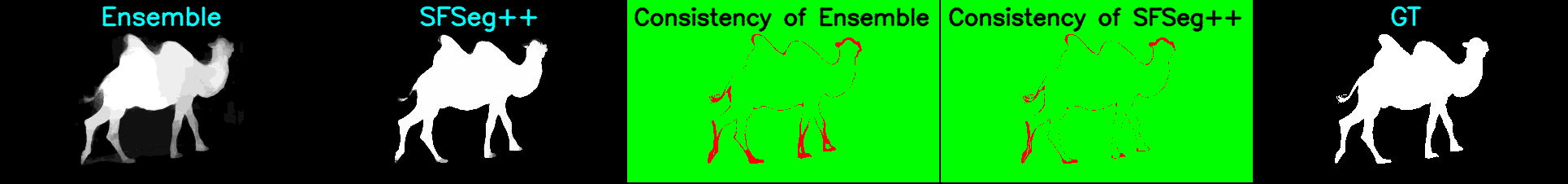}
        \includegraphics[width=0.9\linewidth]{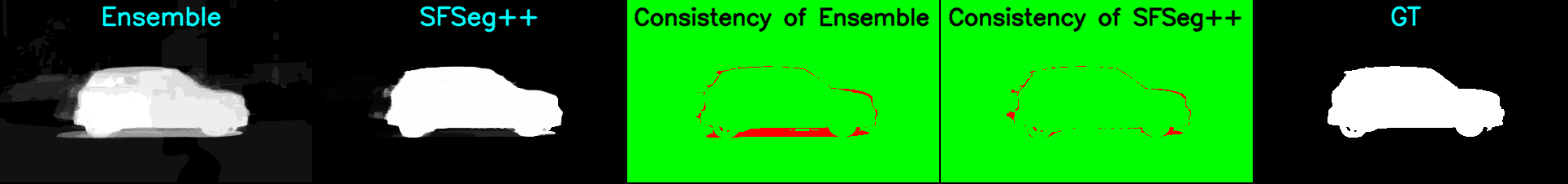}
        \includegraphics[width=0.9\linewidth]{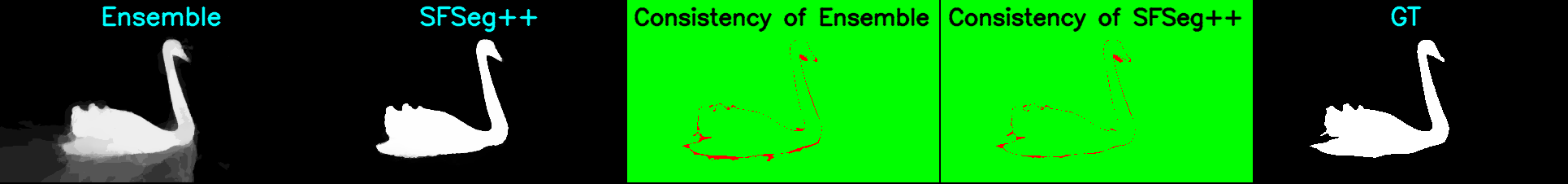}
	\end{center}
	\caption{Visualization examples of temporal consistency: in the first two columns we show the basic Ensemble (average over top $15$ methods) and our SFSeg++. \color{myblue} In the 3$^{rd}$ and 4$^{th}$ column we show the distance between the predicted mask used in TCONT metric (for the Ensemble and SFSeg++ respectively) and ground-truth. Green represents regions which are temporal consistent, while with red we see inconsistent regions. The ground-truth is shown in the last column. TCONT views are computed using Eq.~\ref{eq: temp_consist_output}, as explained in Sec.~\ref{subsec: temporal_consistency}, taking into account a temporal window (left, middle, and right frame) averaged after warping with optical flow. Please observe that the average map, used for measuring consistency, is crisper and of better quality in the case of SFSeg++. Larger green regions for SFSeg++ show on one hand, a better consistency with respect to itself and with respect to ground-truth, given the independently computed object motion field (as estimated by the optical flow).\color{black} Best seen in color.}
	\label{fig: temporal_consist}
\end{figure*}

\begin{table}
\caption{SFSeg++ performance and comparisons to top methods and ensembles over DAVIS-2016 validation set. When starting from 15 input methods, both SFSeg and SFSeg++ improve over their input, outperforming competition, which comprises of state-of-the-art methods on DAVIS-2016 (+0.3\% vs +4.2\%) and the classic Ensemble, as average over its input channels (+0.1\% vs +3.1\%). Most powerful, SFSeg++ proves to bring complementary value to supervised learning, as it outperforms the best learned NN Ensemble by 1.1\%.}
\begin{center}
	\begin{tabular}{lrrr}
		\toprule
		\multicolumn{1}{l}{Method} &
		\multicolumn{1}{p{1.cm}}{\raggedleft Jaccard (\%)} &
		\multicolumn{1}{p{1.3cm}}{\raggedleft J Diff over\\Best (\%)} &
		\multicolumn{1}{p{1.9cm}}{\raggedleft J Diff over\\Ensemble (\%)} \\
        \midrule
		Weakest Method (TRC~\cite{trc}) & 47.3 & -35.1 & -36.2 \\
		$3^{rd}$ Best Method (COSNet~\cite{cosnet}) & 80.5 & -1.9 & -3.0\\
		$2^{nd}$ Best Method (AnDiff~\cite{andiff}) & 81.7 & -0.7 & -1.5\\
		Best Method (MATNet~\cite{matnet}) & 82.4 & 0 & -1.1 \\
		SFSeg(Best Method)  & 82.7 &  +0.3 & -0.8 \\
        Ensemble  & 83.5 & +1.1 & 0\\
		SFSeg(Ensemble) & 83.6 &  +1.2 & +0.1 \\
		Best NN Ensemble & 85.5 & +3.1 & +2.0 \\
		\textbf{SFSeg++} & \textbf{86.6} & \textbf{+4.2} & \textbf{+3.1} \\
		\bottomrule
    \end{tabular}
\end{center}
\label{tab: main_learning_exp}
\end{table}

\subsubsection{\textbf{TCONT: Temporal consistency}}
\label{subsec: exp_temporal_consistency}
We analyze next the influence of our SFSeg++ algorithm on the temporal consistency of the segmentation masks over the video volume, using the newly introduced TCONT metric, detailed in Sec.~\ref{subsec: temporal_consistency}.
We apply TCONT with respect to ground-truth, for the basic Ensemble and SFSeg++, showing the results in Tab.~\ref{tab: consistency_exp}. We also present some qualitative samples focusing on the temporal consistency in Fig.~\ref{fig: temporal_consist} (green regions are temporal consistent).

\noindent \textbf{Key observations:} After applying SFSeg++, the temporal consistency of the output is significantly improved by 4.5\% with respect to ground-truth. This highlights the fact that our algorithm not only improves the overall segmentation accuracy, but also makes it more consistent with the object motion field, a property that is desired in the context of videos. We believe that the main reason for our improved temporal consistency comes directly from the fact that our special video 3D filtering is designed from the start having both space and time in mind, which makes it more suitable for object segmentation in video than other more traditional approaches that start from the level of individual frames.

Intuitively, the difference between two methods having a similar frame-level accuracy, but with different consistency measures at temporal, video level, could be better understood by the following example. Even though the two approaches have, initially similar accuracy, the method which is more stable with respect to the motion field is more likely to produce a segmentation of higher quality after optical flow warping than the one which is less stable with respect to motion. Therefore by measuring such stability with respect to motion, we are in fact looking at the ability of a method to intrinsically consider time and motion, and this is definitely a desired property for any video object segmentation procedure.\\

\begin{table}[t]
\caption{TCONT metric for measuring temporal consistency. We compute the TCONT metric for best and weakest methods, for the Ensemble over all 15 input methods and for our SFSeg++. SFSeg++ increases the temporal consistency of the input by 4.5\% with respect to GT when compared with the basic Ensemble.}
\label{tab: consistency_exp}
\begin{center}
	\begin{tabular}{r c}
		\toprule
		& \multicolumn{1}{c}{\textbf{Temporal Consistency}}\\
		& IoU wrt GT (\%)\\
        \midrule
        Weakest Method \cite{trc} & 46.0 \\
        Best Method \cite{matnet} & 77.0 \\
        \cmidrule(lr){1-2}
        Ensemble & 75.7\\
        \textbf{SFSeg++}  & \textbf{80.2} \\
        \cmidrule(lr){2-2}
                 & \textbf{+4.5\%} \\
		\bottomrule
    \end{tabular}
\end{center}
\end{table}

\subsubsection{\textbf{Running time}} The SFSeg algorithm scales well, its runtime being linear in the number of video pixels, as detailed in Sec.~\ref{sec: algorithm}. For a frame of $480\times854$ pixels, it takes $0.01$ sec per iteration, compared with $0.8$ sec for denseCRF. The time penalty of adding SFSeg is minor for most methods, which take several seconds per frame (\eg $4.5$ sec per frame \cite{osvoss}, $13$ sec per frame \cite{premvos}). We tested on a GTX Titan X Maxwell GPU, in Pytorch~\cite{pytorch}.

\begingroup
\setlength{\tabcolsep}{5pt} 
\begin{table*}[t]
    \caption{Comparison on 5 tracking benchmarks. In the first group, we show individual methods, used as input for our SFTrack++. In the second one, we show ensemble methods: a basic (median) and a neural net model with a similar number of parameters like SFTrack++. Our method outperforms both the input or other ensemble methods by a large margin on the challenging benchmarks GOT-10k and TrackingNet, while obtaining competitive results on OTB, UAV, and NFS. For SFTrack++ we report mean and std when training the model from scratch three times. With blue, we represent the best single method in the column, and with red the best ensemble. The raw results are available in the supplementary material.}
    \begin{center}
        \begin{tabular}{l| r c  c  c ccc  ccc}
        \toprule
        & \multicolumn{1}{c}{\textbf{Method}} &
        \multicolumn{1}{c}{\textbf{OTB}} &
        \multicolumn{1}{c}{\textbf{UAV}} &
        \multicolumn{1}{c}{\textbf{NFS}} &
        \multicolumn{3}{c}{\textbf{GOT-10k}} &
        \multicolumn{3}{c}{\textbf{TrackingNet}} \\
        & & AUC & AUC & AUC & AO & SR$_{50}$  & SR$_{75}$ & Prec & Prec$_{norm}$ & AUC\\
        \cmidrule(l){1-1}
        \cmidrule(l){2-2}
        \cmidrule(l){3-3}
        \cmidrule(l){4-5}
        \cmidrule(l){6-8}
        \cmidrule(l){9-11}
        \multirow{5}{*}{\rotatebox[origin=c]{90}{Single Method}}
        & D3S  &  57.7 & 45.0 & 38.6 &
        39.3 & 39.0 & 10.1 &
        52.2 & 67.9 & 52.4\\
        & SiamBAN   &  67.6 & 60.8 & 54.2 & 
        54.6 & 64.6 & 40.5 & 
        68.4 & 79.5 & 72.0 \\
        & ATOM-18    &  66.7 & 64.3 & 58.4 & 
        55.0 & 62.6 & 39.6 & 
        64.8 & 77.1 & 70.3\\
        & SiamRPN++ &  65.0 & \textbf{\color{red}65.0} & 50.0 & 
        51.7 & 61.5 & 32.5 & 
        \textbf{\color{red}69.3} & 80.0 & 73.0\\
        & PrDimp-18  &  \textbf{\color{red}67.6} & 63.5 & \textbf{\color{red}62.6} &
        \textbf{\color{red}60.8} & \textbf{\color{red}71.0} & \textbf{\color{red}50.3} & 
        69.1 & \textbf{\color{red}80.3} & \textbf{\color{red}75.0} \\
        \cmidrule(l){1-1}
        \cmidrule(l){2-2}
        \cmidrule(l){3-3}
        \cmidrule(l){4-5}
        \cmidrule(l){6-8}
        \cmidrule(l){9-11}
        \multirow{4}{*}{\rotatebox[origin=c]{90}{Ensemble}}
        & Basic (median) & 66.6 & 60.8 & 55.5 & 
        54.7 & 63.9 & 31.6 &
        69.0 & 80.0 & 73.9\\
        & Neural Net & \textbf{\color{myblue}71.3} & 59.7 & 58.2 & 
        59.5 & 69.8 & 42.9 &
        70.6 & 80.2 & 74.5 \\
        & \textbf{SFTrack++} & 70.3 & \textbf{\color{myblue}61.2} & \textbf{\color{myblue}62.4} & 
        \textbf{\color{myblue}62.0} & \textbf{\color{myblue}73.3} & \textbf{\color{myblue}47.8} & 
        \textbf{\color{myblue}71.9} & \textbf{\color{myblue}81.9} & \textbf{\color{myblue}76.1}\\
        & std & $\pm$ \small 0.5 & $\pm$ \small 0.2 & $\pm$ \small 0.1 & $\pm$ \small 0.7 & $\pm$ \small 0.5 & $\pm$ \small	1.1& $\pm$ \small	0.3 & $\pm$ \small	0.3 & $\pm$ \small	1.0\\
        \bottomrule
        \end{tabular}
    \end{center}
    \label{tab:baselines}
\end{table*}
\endgroup

\subsection{Spectral Segmentation integrated in Tracking}
\label{subsec: exp_sftrack++}
We test if SFTrack++ brings in a complementary dimension to object tracking by having an intermediary fine-grained representation, extracted over multiple state-of-the-art trackers' outputs, and smoothed in space and time. We guide our experiments such that we evaluate the least expensive pathways first. For reducing the hyper-parameters search burden, we use AdamW~\cite{adamw-amsgrad}, with a scheduler policy that reduces the learning rate on a plateau. For efficiency and compactness, we use the same channels to construct both the unary and pairwise maps: $\mathbf{S}_i = \mathbf{F}_i$. Their learned weights are also shared $w_{s, i} = w_{f, i}$. We use as input channels bounding boxes extracted with top single object trackers. 

For each input tracker method considered, we run it in advance on all benchmarks (on training, valid, and test splits) to generate pre-processed input. We also generate the ground-truth bounding box segmentations for all benchmarks. This speeds up our training, making the overall training and testing self-contained,
independent w.r.t. the input methods' code. We resized each frame to keep its aspect ratio, having its maximum dimension of $480$ pixels. For training, for each video in the tracking benchmarks, we used only a sample with 5 frames. The pre-processed data for one single tracker, on segmentation and tracking benchmarks, for training, valid, and testing splits takes $\approx1$ TB.  We choose $5$ trackers (PrDiMP~\cite{prdimp}, ATOM~\cite{atom}, D3S~\cite{d3s}, SiamRPN++~\cite{siamrpn++}, SiamBAN~\cite{siamban}), which differ in architecture, training sets and overall in their approaches, but all achieves top results on tracking benchmarks. We integrate them next in the PyTracking~\cite{pytracking} framework. For extracting the bounding box coordinates out of the segmentation mask ($NN_{seg2box}$) we use in all our experiments the region proposal from scikit-learn~\cite{scikit-learn}, with a $0.75$ threshold for binarization.\\

\noindent\textbf{Training.} In Phase 1 we train our $NN_{feat2seg}$ network on DAVIS-2017~\cite{davis-2017} and Youtube-VIS~\cite{Youtube-VIS} trainsets, for each individual object. It receives the current RGB and the output of a tracking method (bbox), randomly sampled at training time. We use the U-Net architecture, validating the right number of parameters ($100$K - $1$ mil) and the number of layers. We use DAVIS-2017 and Youtube-VIS evaluation sets to stop the training. Following the curriculum learning approach, before introducing tracking methods into the pipeline, we use at the beginning of the tracking GT bounding boxes (extracted from segmentation GT, as straight bounding boxes). This allows the $NN_{feat2seg}$ component to get a good initialization, before introducing faulty bounding box extractors, namely the top 5 tracking methods mentioned before. For Phase 2, we also train for the segmentation task. We learn the second part of our method to have an intermediary fine-grained representation, extracted over multiple channels, and smoothed in space and time. We validate here $N_{iters}$, the number of spectral iterations ($1$-$5$). We train on DAVIS-2016~\cite{davis-2016} and Youtube-VIS datasets. In Phase 3, training for tracking, we learn a regression network, $NN_{seg2box}$ (with $50$K-$500$K parameters), to transform the final segmentation to a bounding box. We train on TrackingNet~\cite{tracking-net} and GOT-10k~\cite{got-10k} training splits.\\

\subsubsection{\textbf{Baselines}}
This comparative experiment focuses on the improvements SFTrack++ could bring over state-of-the-art and other top approaches for object tracking: single-method state-of-the-art trackers, a basic ensemble over trackers, SFTrack++ applied only over the best tracker, SFTrack++ applied over the basic ensemble and the best learned neural net ensemble we could get out of several configurations (2D and 3D versions for U-Net~\cite{unet} and shallow nets, with different number of parameters: $100$K, $500$K, $1$ mil, $5$ mil, $15$ mil). All methods have the same input from top $5$ trackers and train on TrackingNet and GOT-10k train sets as previously described. We evaluate our solution against all baselines on five tracking benchmarks: \textbf{TrackingNet}, \textbf{GOT-10k}, \textbf{NFS}~\cite{nfs}, \textbf{OTB-100}~\cite{otb100} and \textbf{UAV123}~\cite{uav123}. We provide statistical results (mean and variance over several runs) to better indicate a strong positive/negative result, or an inconclusive one.

In Tab.~\ref{tab:baselines} we present the results of our method on five tracking benchmarks: OTB100, UAV123, NFS30, GOT-10k, and TrackingNet, comparing it with other top single methods and ensemble solutions. For single methods, we take into account each input method in our SFTrack++: D3S, SiamBAN, ATOM-18, SiamRPN++, PrDimp-18. We chose only one lightweight configuration per tracker, common across all benchmarks. For ensembles, we use 1) a basic per-pixel median ensemble followed by the same bounding box regressor used in all experiments and we also trained a more complex one: 2) a neural net having an UNet architecture (with $5$ down-scaling and $5$ up-scaling layers) and a similar number of parameters like SFTrack++ ($\approx4.3$ millions). We observe that the variation across different runs (including training from scratch) of our method is very small, showing a robust result and a clear conclusion. On newer, larger, and more generic datasets like GOT-10k and TrackingNet, our method surpasses others by a large margin, while on OTB100, UAV123, and NFS30 it has competitive results.\\

\subsubsection{\textbf{Ablation studies}} We vary several components of our end-to-end model to better understand their role and power. We train our \textbf{Phase 1} component, $NN_{feat2seg}$ net, not only for bounding box input features but also for other earlier features, extracted from each tracker architecture. We test the overall tracking performance for this case.
We remove from the pipeline the spectral refinement in \textbf{Phase 2} and report the results.
We test the performance of our tracker without the \textbf{Phase 3} neural net, $NN_{seg2box}$, by replacing it with a straight box and rotated box extractors from OpenCV~\cite{opencv}.
We test several losses to optimize for both segmentation and tracking tasks: a linear combination between the weighted diceloss~\cite{wdice-loss} and binary cross-entropy, Focal-Tversky~\cite{focal-tversky}, and Focal-Dice~\cite{focal-dice}. For the ablative experiments, we evaluate only on OTB100, UAV123, and NFS30 tracking datasets.

To validate the components of our method, we test in Tab.~\ref{tab:ablations} different variations, reporting results on OTB100, UAV123, and NFS30. First, we remove the spectral refining component from phase 2, taking out the temporal dependency and leaving the per frame predictions independent. In the next experiment, we remove the neural net from phase3 $NN_{segm2bbox}$. In the next chunk, we investigate the number of input methods. Last, we vary the number of spectral iterations from phase 2. The conclusions from this wide ablation are the following: \textbf{a)} the spectral refining component is very important, emphasizing the initial intuition that preserving the object consistency in space and time using our proposed spectral approach improves the overall performance in tracking. \textbf{b)} The quality of the input in our SFTrack++ method is important, but the more methods we use, the better. \textbf{c)} We obtain better results using only one single spectral iteration. 

\subsubsection{\textbf{Qualitative results}} 
\begin{figure}[t]
	\begin{center}
		\includegraphics[width=0.99\linewidth]{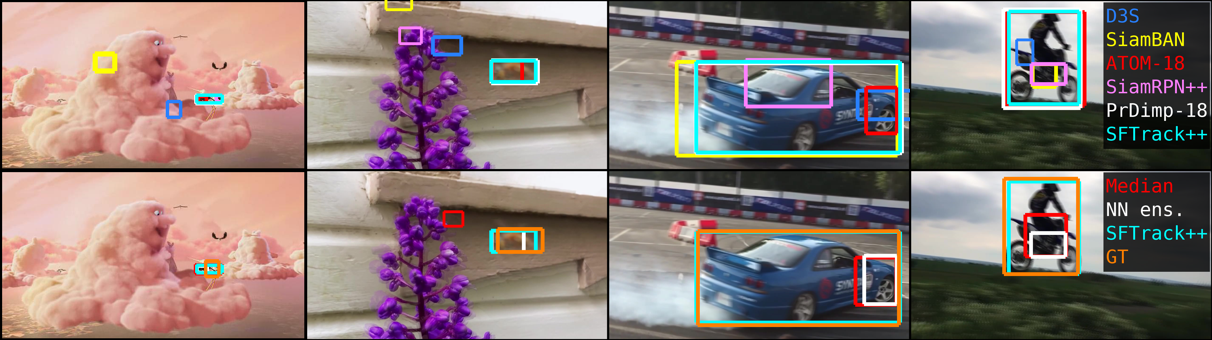}
	\end{center}
	\caption{Qualitative results. We compare in the first line SFTrack++ with the input from individual methods. In the second line, we show the ground truth (in orange) and ensemble methods results that receive the same input as SFTrack++. We notice that even though the other ensembles fail to find a good bounding box, SFTrack++ manages to combine the input methods better, even in cases with a high variance among input methods. Best seen in color.}
	\label{fig: qual_res}
\end{figure}

Since SFTrack++ is an ensemble method, we show in Fig.~\ref{fig: qual_res} difficult cases and how it compares with individual methods and with other ensembles, starting from the same input. We see how our method outperforms the others, even in those hard cases where an agreement seems hard to achieve.

\begingroup
\setlength{\tabcolsep}{3pt} 
\begin{table}[t]
    \caption{Ablations on OTB100+UAV123+NFS30 benchmarks. In the first group, we remove from the pipeline phase 2 and phase 3, respectively. The results show that both components are crucial for the method. Next, we vary the number of input methods (1 to 5) but also their quality (best or median). We see that even the quality of the input matters, using more methods as input improves the overall performance. In the last part, we validate the number of spectral iterations, a single iteration achieving the best score, which slowly degrades over more iterations.}
    \begin{center}
        \begin{tabular}{l cccc}
        \toprule
        \textbf{SFTrack++} variations & \multicolumn{1}{c}{\textbf{OTB}} &
        \multicolumn{1}{c}{\textbf{UAV}} &
        \multicolumn{1}{c}{\textbf{NFS}} & \multicolumn{1}{c}{\textbf{OTB+UAV+NFS}} \\
        \cmidrule(l){1-1}
        \cmidrule(l){2-5}
        w/o Spectral Refinement (phase two) & \textbf{71.6} & 60.5 & 60.9 & 64.0\\
        w/o NN$_{segm2bbox}$ (phase three) & 65.5 & 57.4 & 58.5 & 60.2\\
        \cmidrule(l){1-1}
        \cmidrule(l){2-5}
        Median (over 5 methods) as input & 70.8 & 60.8 & 60.0 & 63.7\\
        Best method (PrDimp-18) as input & 67.1 & 59.7 & 61.3 & 62.5\\
        Top 3 methods as input & 64.8 & 60.8 & 61.1 & 62.1\\
        \cmidrule(l){1-1}
        \cmidrule(l){2-5}
        2 spectral iterations & 70.3 & 60.9 & 61.8 & 64.1\\
        3 spectral iterations & 68.0 & 61.0 & 60.1 & 62.9\\
        \cmidrule(l){1-1}
        \cmidrule(l){2-5}
        \textbf{SFTrack++} (1 iter, 5 methods) & 70.3 & \textbf{61.2} & \textbf{62.4} & \textbf{64.5}\\
        \bottomrule
        \end{tabular}
    \end{center}
    \label{tab:ablations}
\end{table}
\endgroup

\color{black}
\section{Concluding remarks}
\label{sec: conclusions}
We formulate video object segmentation as clustering in the space-time graph of pixels. We first introduce an efficient spectral algorithm, Spectral Filtering Segmentation (\textbf{SFSeg}), in which the standard power iteration for computing the principal eigenvector of the graph's adjacency matrix is transformed into a set of special 3D convolutions applied on 3D feature maps in the video volume. Our original theoretical contribution makes the initial intractable problem possible. Then we extend this approach into \textbf{SFSeg++}, which fully learns to perform efficient spectral clustering with 3D convolutions over multiple of input channels. Further, we go beyond the standalone segmentation task and propose a way to seamlessly integrate our spectral space-time filtering into the tracking task, introducing \textbf{SFTrack++}. This approach proposes a way for exploiting the space-time continuum using an intermediary fine-grained object mask in the context of tracking, as opposed to simple bound box object shape representations.

We validate experimentally the value of each technical contribution. First, we show theoretically that our fast solution based on the first-order Taylor approximation of the original pairwise potential used in spectral clustering is practically equivalent to the original one. Secondly, in experimental comparisons, we show that the single-channel SFSeg consistently improves (for 80\% of videos), when applied as a refinement procedure over all top published video object segmentation methods, at a small additional computational cost. Moreover, we show that SFSeg also achieves top performance in combination with other backbone networks (not necessarily state-of-the-art). Thirdly, we demonstrate the effectiveness and robustness of learning over multi-channels by experimenting with different types of noisy input channels. Then we introduce \textbf{TCONT}, a temporal consistency measure, which we use to analyze the methods' consistency with respect to the object motion field in video. The analysis indicates that our formulation is indeed more stable with respect to the object's motion field (computed independently), which validates our motivation for the spatio-temporal clustering approach that is meant to preserve such stability in space and time. And last, we show how our spectral segmentation formulation (SFSeg++) can be seamlessly extended to become a top-quality tracking system (SFTrack++). We achieve state-of-the-art results with both segmentation and tracking approaches (SFSeg++ and SFTrack++, respectively), on multiple competitive benchmarks.

In a nutshell, we validate all four key aspects of our contributions: $\mathbf{1)}$ the efficiency of the 3D filtering approach as a way to perform spectral space-time clustering; $\mathbf{2)}$ the power of learning over multiple channels and the algorithm's robustness to noise; $\mathbf{3)}$ the superior temporal consistency of our approach is superior, measured with newly introduced metric; $\mathbf{4)}$ the value of the approach beyond the bounds of segmentation, on object tracking, with state of the art results.

As a last note, it is important that our spectral clustering with minimal number of parameters for learning is better and more robust in the face of limited supervised training data, than powerful ensembles learned with deep neural networks. The consistent improvements in practice over different top-quality input channels, brought by SFSeg, SFSeg++, as well as SFTrack++, indicate that our spectral approach brings a new and complementary dimension to clustering in space-time, as needed in important visual tasks, such as object segmentation and tracking in video.\\

\textbf{Acknowledgments:} Work funded in part by UEFISCDI, under Projects EEA-RO-2018-0496 and PN-III-P4-ID-PCE-2020-2819.

\bibliographystyle{elsarticle-num}
\bibliography{ai}




%

\end{document}